\documentclass[runningheads]{llncs}
\usepackage[T1]{fontenc}
\usepackage{graphicx}
\usepackage{hyperref}
\usepackage{color}

\usepackage{tabularx}
\usepackage{verbatim}
\usepackage{tikz}
\usepackage{pgfplots} 
\pgfplotsset{compat=1.18}
\usetikzlibrary{patterns}
\usetikzlibrary{positioning,decorations.pathreplacing,calligraphy}
\usepackage{csvsimple-l3}
\usepackage{bbm} 

\usepackage{ifthen}
\newcommand{\domain}[2][]{\ensuremath{\mathcal{#2}\ifthenelse{\equal{#1}{}}{}{^{\left(#1\right)}}}}
\newcommand{\prob}[2][]{\ensuremath{\mathbb P\ifthenelse{\equal{#1}{}}{}{^{\left(#1\right)}}_{#2}}}

\usepackage{float}
\usepackage{stfloats}
\usepackage{pifont}
\newcommand{\cmark}{\textcolor{green!80!black}{\ding{51}}} 
\newcommand{\xmark}{\textcolor{red}{\ding{55}}} 
\usepackage{xspace}
\usepackage{booktabs}

\usepackage{amsmath,amsfonts,amssymb}

\usepackage[capitalise]{cleveref}
\usepackage{subcaption}
\Crefname{equation}{Eq.}{Eqs.}
\Crefname{figure}{Fig.}{Figs.}
\Crefname{table}{Tab.}{Tabs.}
\Crefname{section}{Sec.}{Secs.}

\begin{document}

\newcommand{\eg}{\emph{e.g.}\xspace}
\newcommand{\ie}{\emph{i.e.}\xspace}
\newcommand{\cf}{\emph{cf.}\xspace}

\title{Evaluation of Randomization through Style Transfer for Enhanced Domain Generalization}
\titlerunning{Evaluation of Randomization through Style Transfer}

\author{Dustin Eisenhardt\inst{1,2,3}\orcidID{0009-0007-6777-9038} \and\\ Timothy Schaumlöffel\inst{3,4}\orcidID{0009-0007-4748-3037} \and\\ Alperen Kantarci\inst{3}\orcidID{0000-0002-4080-5538} \and\\ Gemma Roig\inst{3,4}\orcidID{0000-0002-6439-8076}}



\institute{
    German Cancer Research Center (DKFZ), Heidelberg, Germany\and
    German Cancer Consortium (DKTK), partner site Frankfurt/Mainz, a partnership between DKFZ and UCT Frankfurt-Marburg, Frankfurt am Main, Germany\and
    Goethe University Frankfurt, Department of Computer Science, Frankfurt am Main, Germany\and
    The Hessian Center for Artificial Intelligence (hessian.AI), Darmstadt, Germany\\
    \email{dustin.eisenhardt@dkfz-heidelberg.de}
}
\maketitle              

\begin{abstract}
Deep learning models for computer vision often suffer from poor generalization when deployed in real-world settings, especially when trained on synthetic data due to the well-known Sim2Real gap.
Despite the growing popularity of style transfer as a data augmentation strategy for domain generalization, the literature contains unresolved contradictions regarding three key design axes: the diversity of the style pool, the role of texture complexity, and the choice of style source.
We present a systematic empirical study that isolates and evaluates each of these factors for driving scene understanding, resolving inconsistencies in prior work.
Our findings show that (i) expanding the style pool yields larger gains than repeated augmentation with few styles, (ii) texture complexity has no significant effect when the pool is sufficiently large, and (iii) diverse artistic styles outperform domain-aligned alternatives. 
Guided by these insights, we derive \emph{StyleMixDG} (Style-Mixing for Domain Generalization), a lightweight, model-agnostic augmentation recipe that requires no architectural modifications or additional losses.
Evaluated on the GTAV $\to$ \{BDD100k, Cityscapes, Mapillary Vistas\} benchmark, StyleMixDG demonstrates consistent improvements over strong baselines, confirming that the empirically identified design principles translate into practical gains. The code will be released on GitHub.
\end{abstract}

\begin{figure}[t]
    \begin{center}\begin{tikzpicture}
        \node at (2,0) {Dataset Pre-Processing};

        \draw [very thick] (6,0) -- (6, -6);

        \node at (9,0) {Training};

        \foreach \offsetcount/\imagename in {
          1/218, 2/6426, 3/8984
        } \node [name=std\offsetcount] at (0 + 0.1*\offsetcount, -1 -0.1*\offsetcount) {\includegraphics[width=1cm,height=1cm]{Resources/gtr/\imagename}};
        \node[below of = std2,align=center] {Style\\Dataset};

        \node [name=simg11, right of = std2, xshift=-0.9cm + 1.5cm + 0.1cm, yshift=-0.65cm - 0.1cm] {\includegraphics[width=1cm,height=1cm]{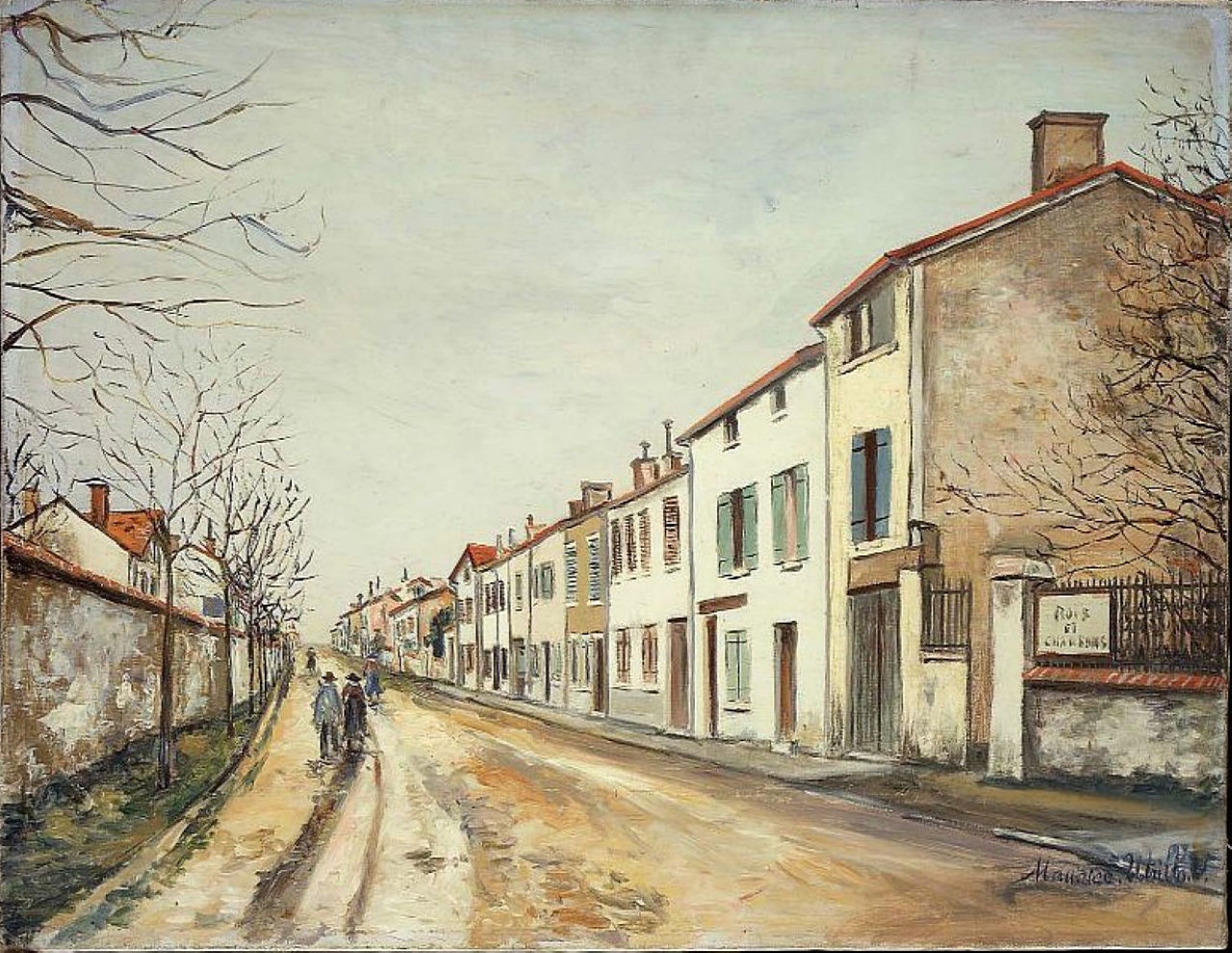}};
        \node [name=simg12, right of = std2, xshift=-0.9cm + 1.5cm + 0.2cm, yshift=-0.65cm - 0.2cm] {\includegraphics[width=1cm,height=1cm]{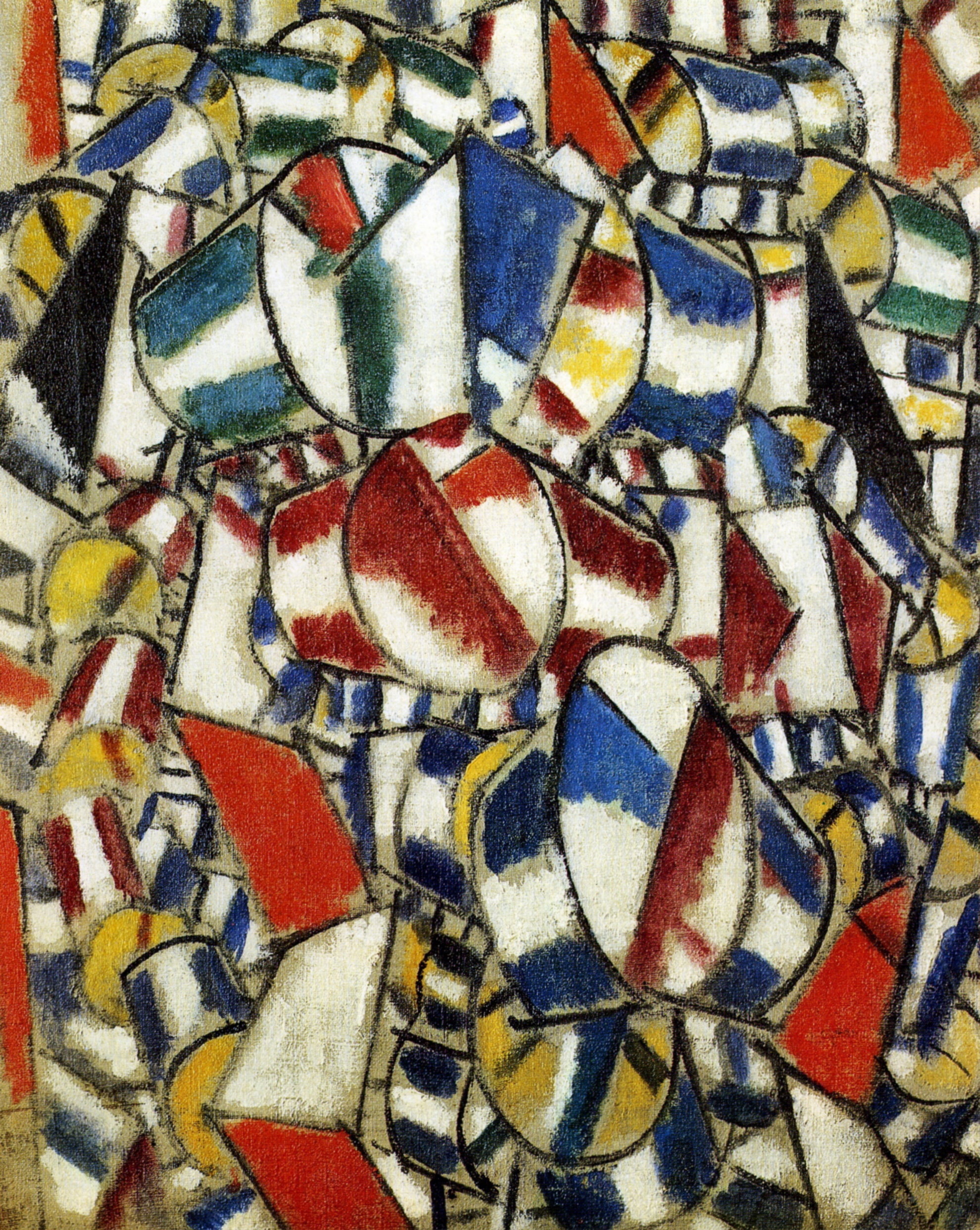}};
        \node [name=simg13, right of = std2, xshift=-0.9cm + 1.5cm + 0.3cm, yshift=-0.65cm - 0.3cm] {\includegraphics[width=1cm,height=1cm]{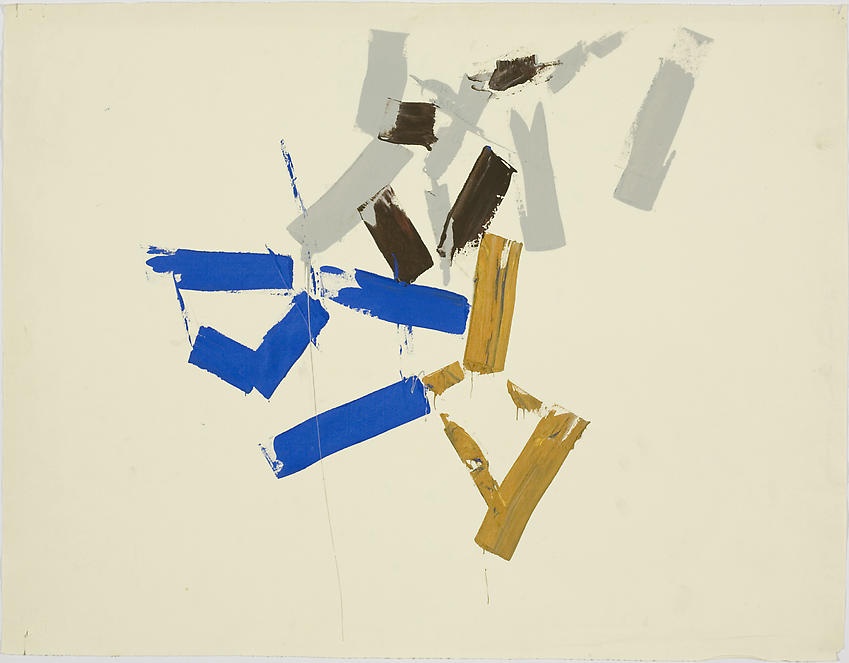}};
        \draw [->] (std2.east) to [bend left] (simg12.north west);

        \node [name=simg21, right of = std2, xshift=-0.9cm + 3.0cm + 0.1cm, yshift=-0.65cm - 0.1cm] {\includegraphics[width=1cm,height=1cm]{Resources/tcps/19492}};
        \node [name=simg22, right of = std2, xshift=-0.9cm + 3.0cm + 0.2cm, yshift=-0.65cm - 0.2cm] {\includegraphics[width=1cm,height=1cm]{Resources/tcps/7129}};
        \node [name=simg23, right of = std2, xshift=-0.9cm + 3.0cm + 0.3cm, yshift=-0.65cm - 0.3cm] {\includegraphics[width=1cm,height=1cm]{Resources/tcps/12271}};
        \draw [->] (std2.east) to [bend left] (simg22.north west);

        \node [name=simg31, right of = std2, xshift=-0.9cm + 4.5cm + 0.1cm, yshift=-0.65cm - 0.1cm] {\includegraphics[width=1cm,height=1cm]{Resources/tcps/12271}};
        \node [name=simg32, right of = std2, xshift=-0.9cm + 4.5cm + 0.2cm, yshift=-0.65cm - 0.2cm] {\includegraphics[width=1cm,height=1cm]{Resources/tcps/19492}};
        \node [name=simg33, right of = std2, xshift=-0.9cm + 4.5cm + 0.3cm, yshift=-0.65cm - 0.3cm] {\includegraphics[width=1cm,height=1cm]{Resources/tcps/7129}};
        \draw [->] (std2.east) to [bend left] (simg32.north west);



        \node [name=aimg11, below of = simg11, yshift=-1.4cm] {\includegraphics[width=1cm,height=1cm]{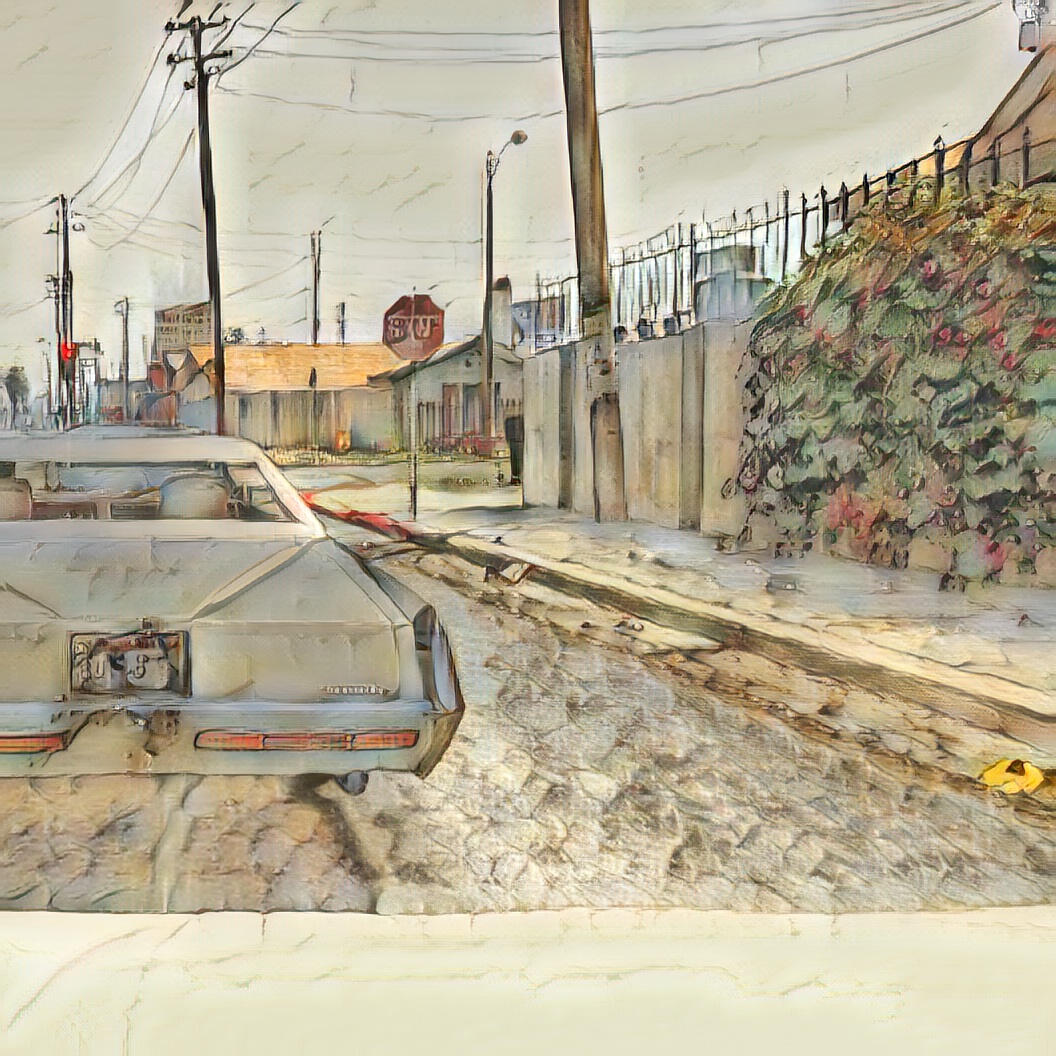}};
        \node [name=aimg12, below of = simg12, yshift=-1.4cm] {\includegraphics[width=1cm,height=1cm]{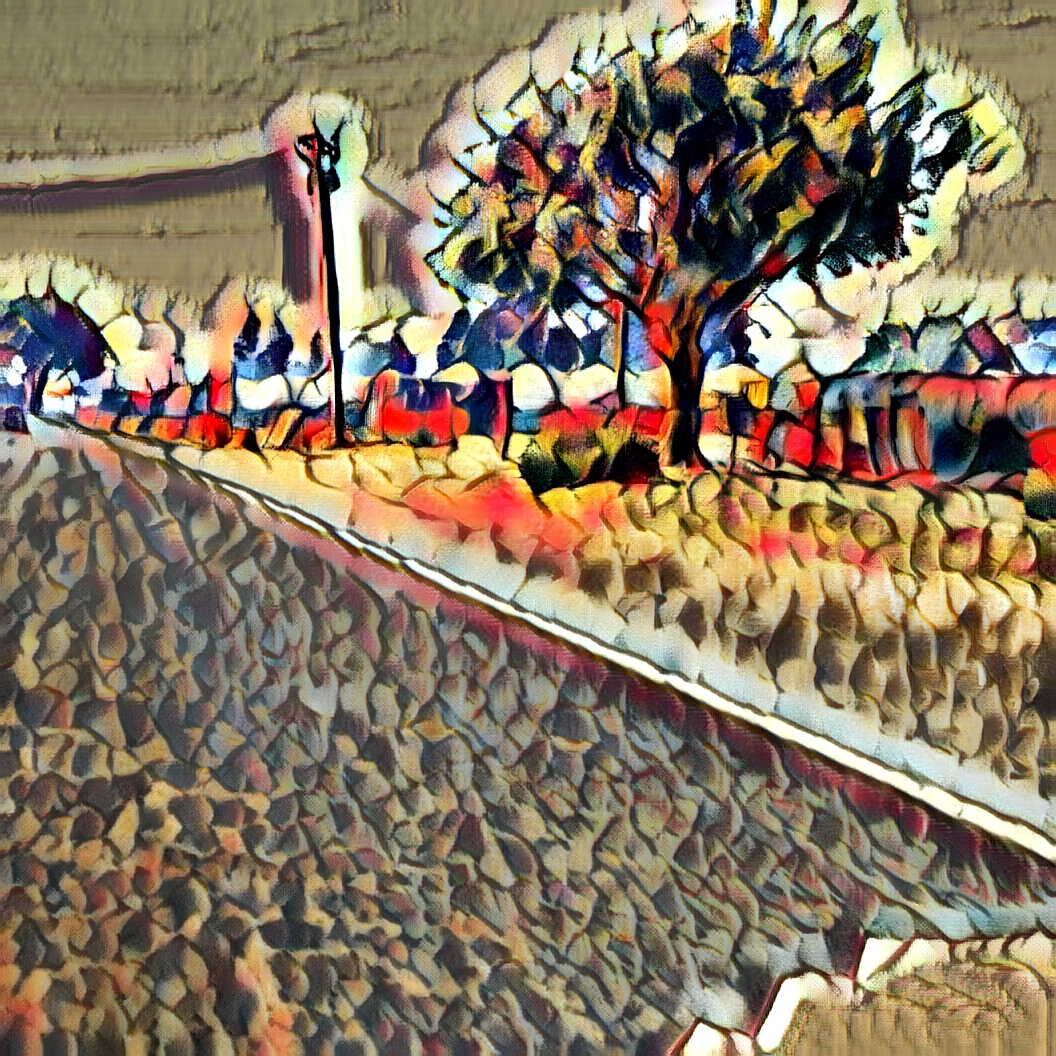}};
        \node [name=aimg13, below of = simg13, yshift=-1.4cm] {\includegraphics[width=1cm,height=1cm]{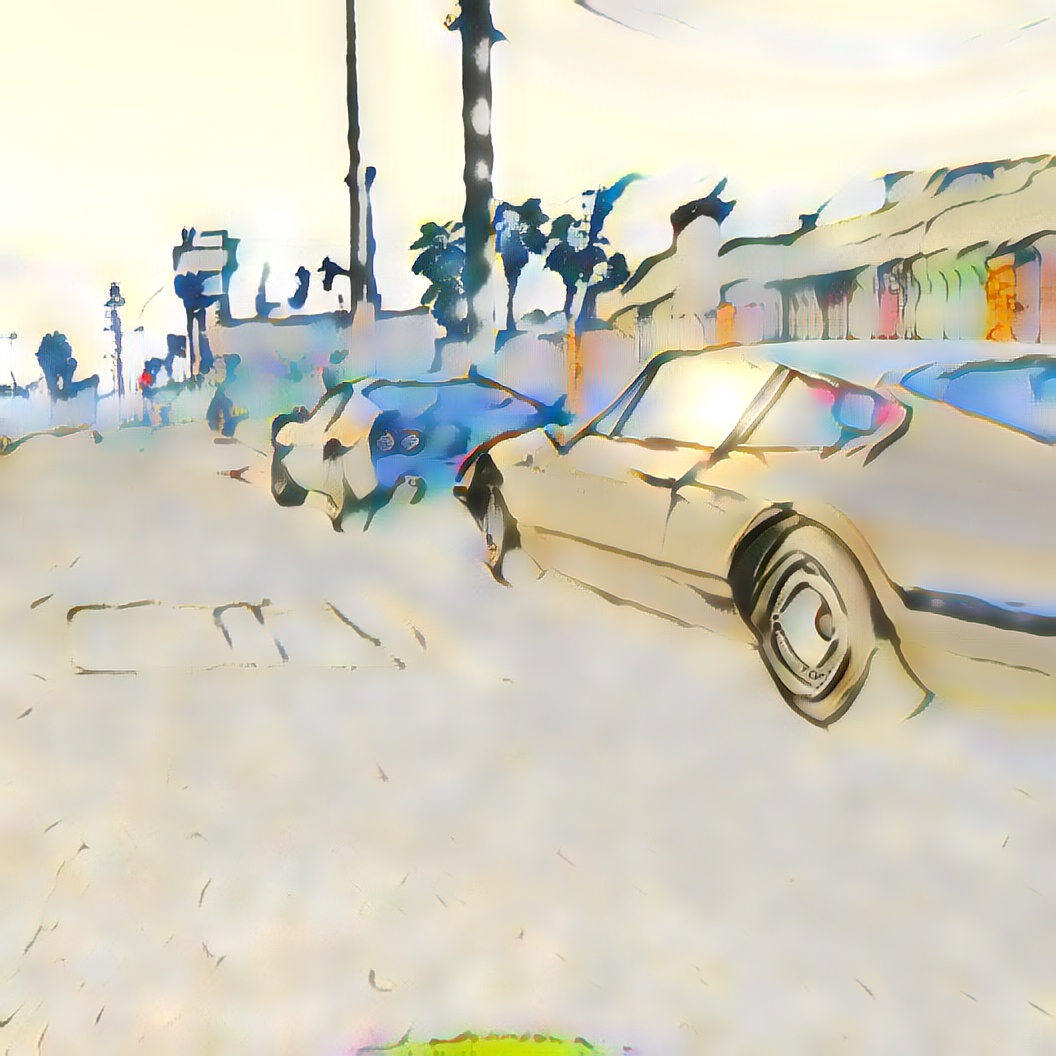}};

        \node [name=aimg21, below of = simg21, yshift=-1.4cm] {\includegraphics[width=1cm,height=1cm]{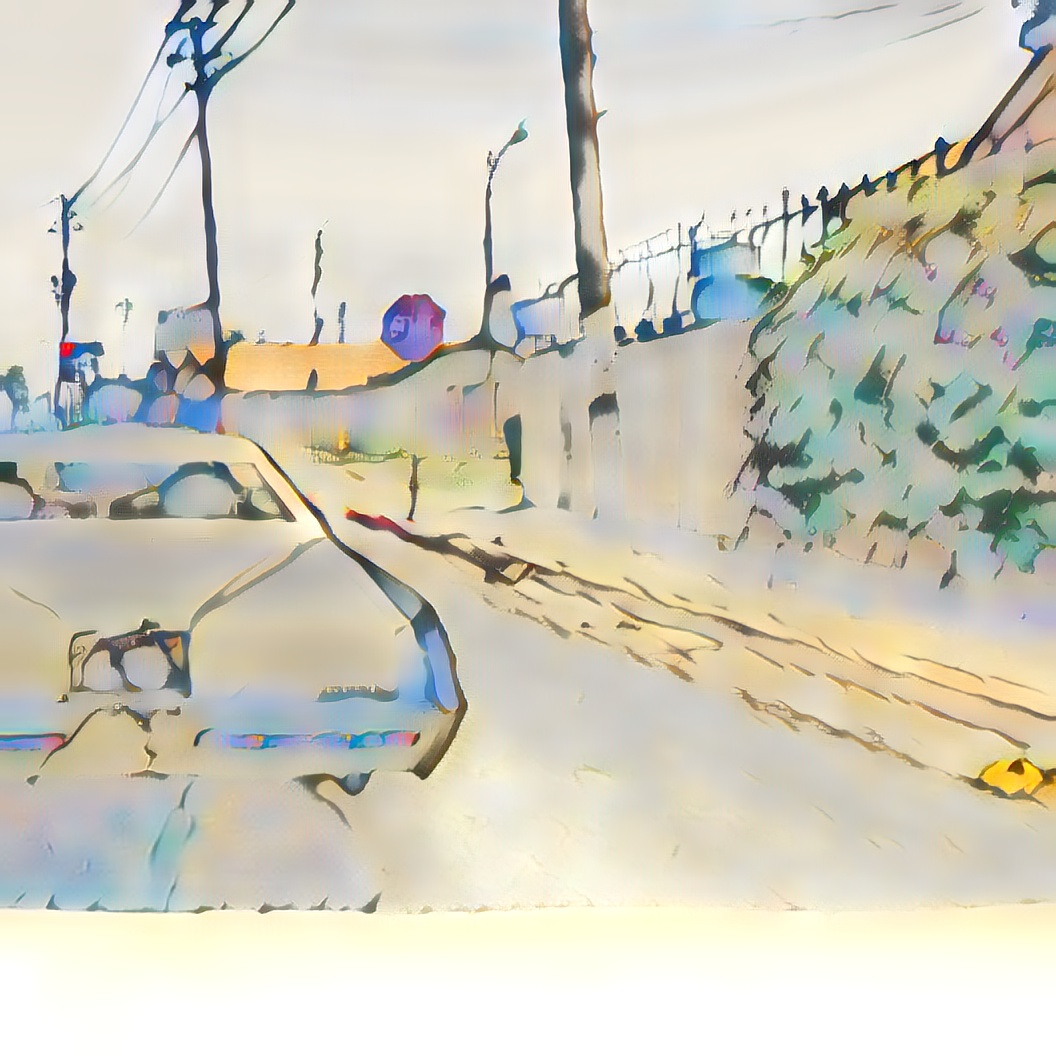}};
        \node [name=aimg22, below of = simg22, yshift=-1.4cm] {\includegraphics[width=1cm,height=1cm]{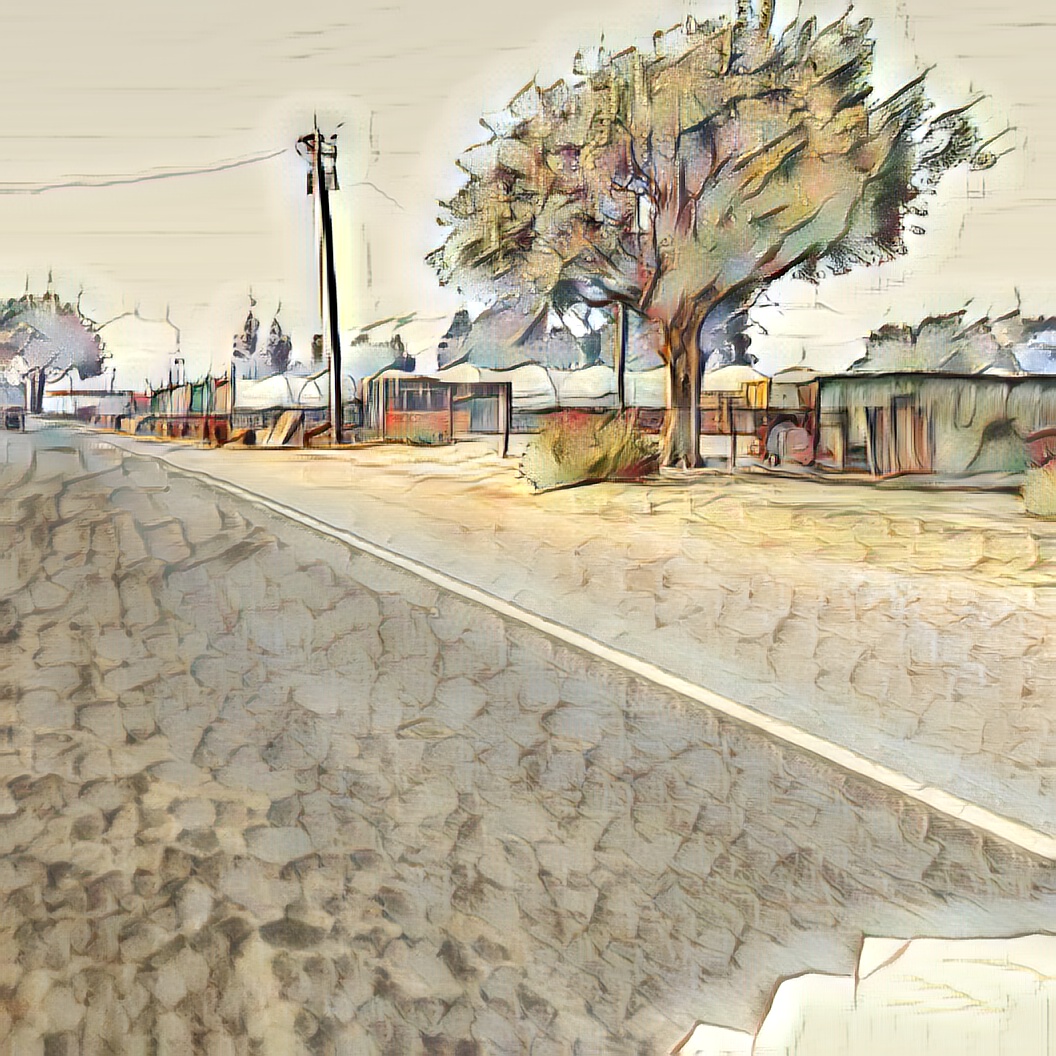}};
        \node [name=aimg23, below of = simg23, yshift=-1.4cm] {\includegraphics[width=1cm,height=1cm]{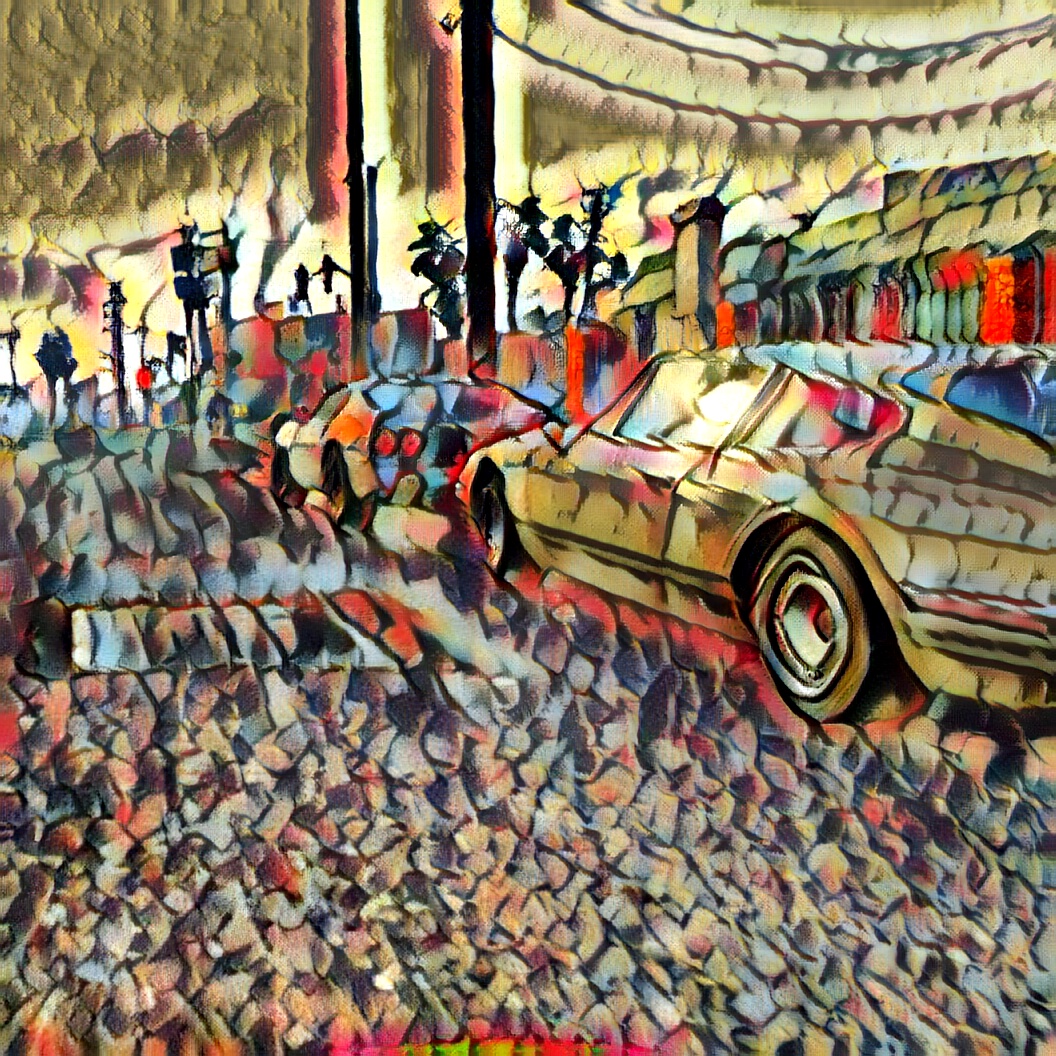}};

        \node [name=aimg31, below of = simg31, yshift=-1.4cm] {\includegraphics[width=1cm,height=1cm]{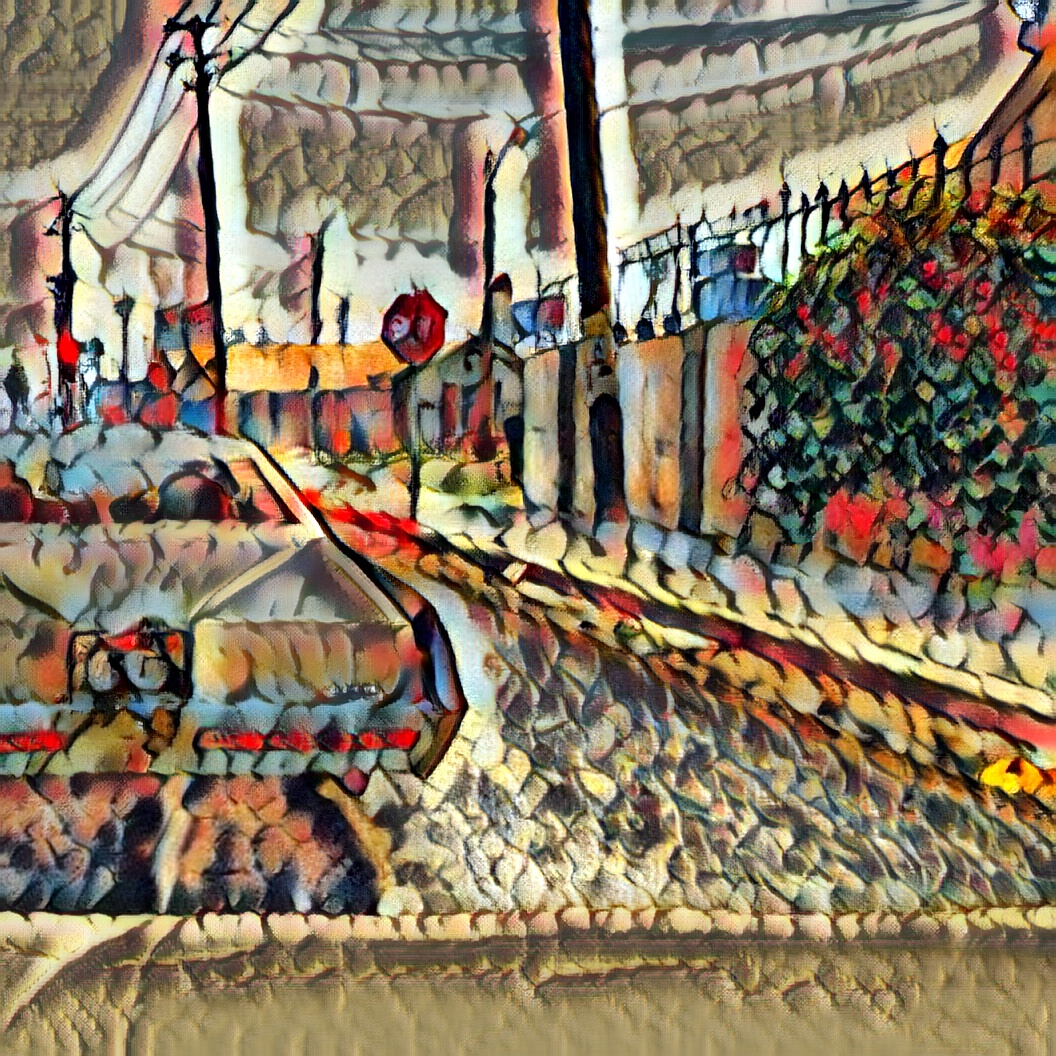}};
        \node [name=aimg32, below of = simg32, yshift=-1.4cm] {\includegraphics[width=1cm,height=1cm]{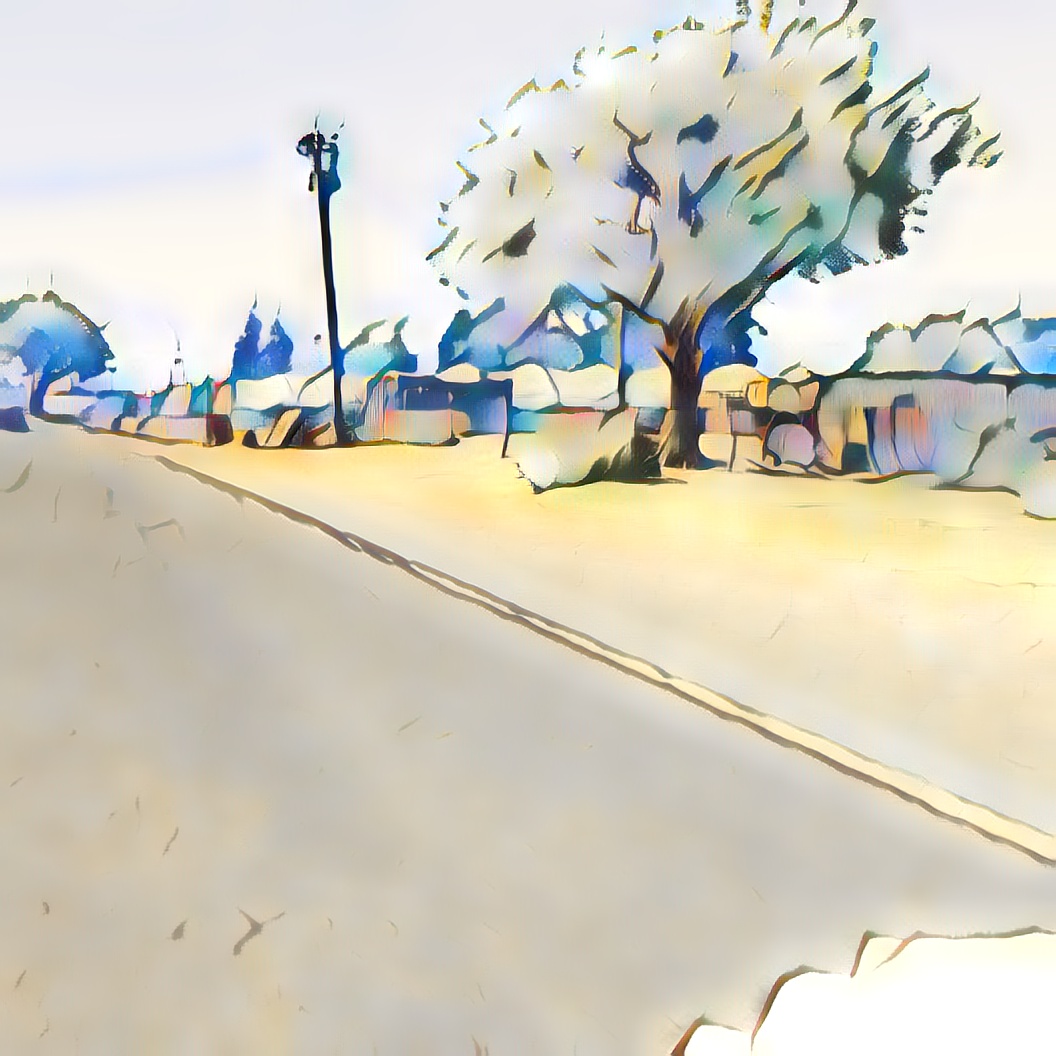}};
        \node [name=aimg33, below of = simg33, yshift=-1.4cm] {\includegraphics[width=1cm,height=1cm]{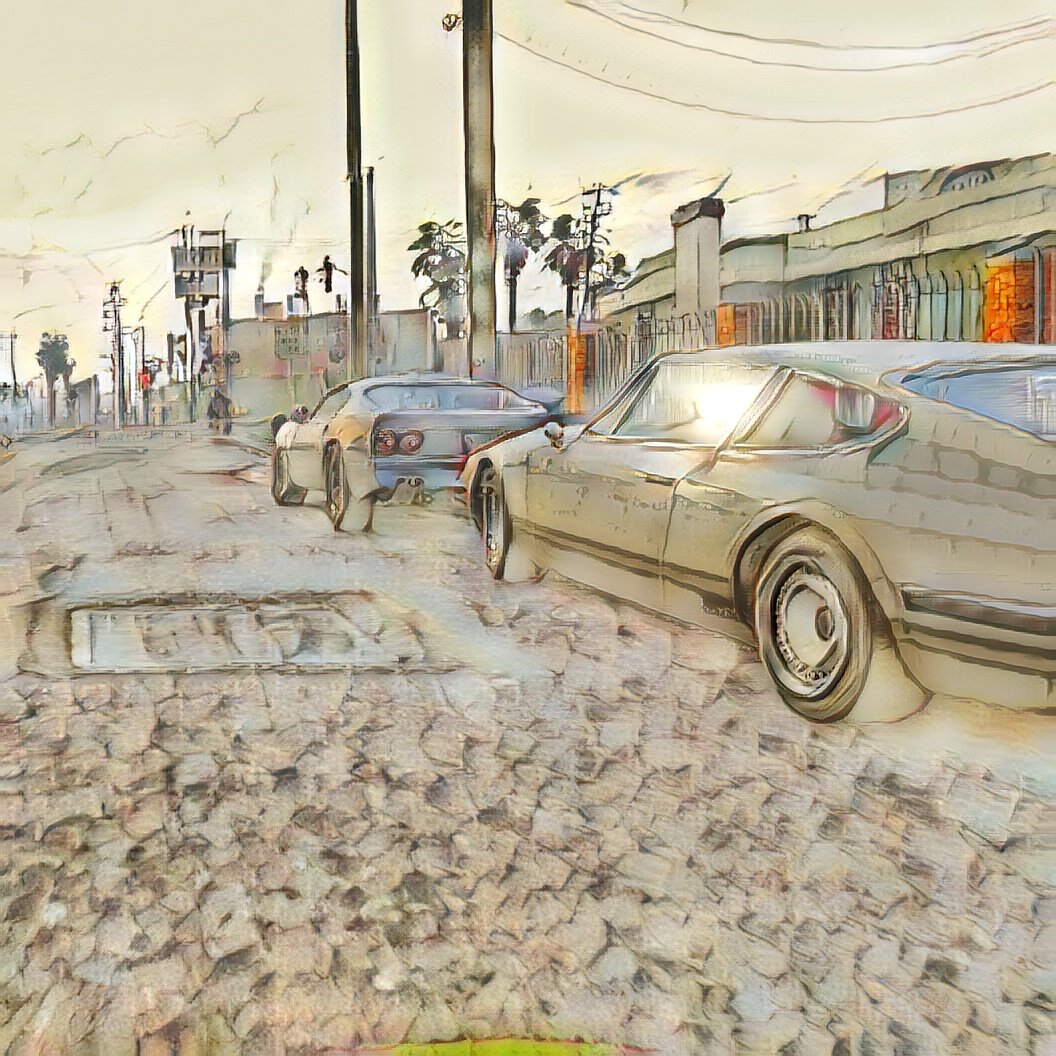}};

        \draw [->] (simg12.south) to (aimg12.north);
        \draw [->] (simg22.south) to (aimg22.north);
        \draw [->] (simg32.south) to (aimg32.north);

        \foreach \offsetcount/\imagename in {
          1/24776_01, 2/00040_01, 3/20078_01
        } \node [name=syd\offsetcount, below of = std2, xshift=-0.2cm + 0.1*\offsetcount cm, yshift=-1.4cm -0.1*\offsetcount cm]  {\includegraphics[width=1cm,height=1cm]{Resources/gtr/\imagename}};
        \node[below of = syd2, align=center] {Synthetic\\Dataset};

        \draw [->] (syd2.east) to [bend left] (aimg12.north west);
        \draw [->] (syd2.east) to [bend left] (aimg22.north west);
        \draw [->] (syd2.east) to [bend left] (aimg32.north west);

        \draw [decorate,decoration={calligraphic brace,mirror},thick] (aimg12.south west) -- (aimg32.south east) node [pos=0.5,below] {$N$-times Augmented Dataset};


        \foreach \offsetcount/\imagename in {
          1/24776_01, 2/00040_01, 3/20078_01
        } \node [name=tsyd\offsetcount, xshift=-0.2cm + 0.1*\offsetcount cm, yshift=-1cm -0.1*\offsetcount cm] at (7, 0) {\includegraphics[width=1cm,height=1cm]{Resources/gtr/\imagename}};


        \node [name=taimg11, right of = tsyd1, xshift=-1.0cm + 1.4cm +0.1cm,yshift=0.1cm + -0.1cm] {\includegraphics[width=1cm,height=1cm]{Resources/gtr/24776_01_7129}};
        \node [name=taimg21, right of = tsyd1, xshift=-1.0cm + 1.4cm +0.2cm,yshift=0.1cm + -0.2cm] {\includegraphics[width=1cm,height=1cm]{Resources/gtr/00040_01_12271}};
        \node [name=taimg31, right of = tsyd1, xshift=-1.0cm + 1.4cm +0.3cm,yshift=0.1cm + -0.3cm] {\includegraphics[width=1cm,height=1cm]{Resources/gtr/20078_01_19492}};

        \node [name=taimg12, right of = tsyd1, xshift=-1.0cm + 2.8cm +0.1cm,yshift=0.1cm + -0.1cm] {\includegraphics[width=1cm,height=1cm]{Resources/gtr/24776_01_19492}};
        \node [name=taimg22, right of = tsyd1, xshift=-1.0cm + 2.8cm +0.2cm,yshift=0.1cm + -0.2cm] {\includegraphics[width=1cm,height=1cm]{Resources/gtr/00040_01_7129}};
        \node [name=taimg32, right of = tsyd1, xshift=-1.0cm + 2.8cm +0.3cm,yshift=0.1cm + -0.3cm] {\includegraphics[width=1cm,height=1cm]{Resources/gtr/20078_01_12271}};

        \node [name=taimg13, right of = tsyd1, xshift=-1.0cm + 4.2cm +0.1cm,yshift=0.1cm + -0.1cm] {\includegraphics[width=1cm,height=1cm]{Resources/gtr/24776_01_12271}};
        \node [name=taimg23, right of = tsyd1, xshift=-1.0cm + 4.2cm +0.2cm,yshift=0.1cm + -0.2cm] {\includegraphics[width=1cm,height=1cm]{Resources/gtr/00040_01_19492}};
        \node [name=taimg33, right of = tsyd1, xshift=-1.0cm + 4.2cm +0.3cm,yshift=0.1cm + -0.3cm] {\includegraphics[width=1cm,height=1cm]{Resources/gtr/20078_01_7129}};

        \draw [decorate,decoration={calligraphic brace,mirror},thick] (tsyd2.south west) -- (tsyd2.south east) node [name=tp,pos=0.5,below] {$20\%$};
        \draw [decorate,decoration={calligraphic brace,mirror,aspect=0.75},thick] (taimg21.south west) -- (taimg23.south east) node [name=ep,pos=0.75,below] {$80\%$};

        \node [name=rc,draw,rectangle,rounded corners] at (9, -3.5) {Weighted Random Choice};
        \draw [->] (tp.south) -- ++(0,-0.5cm);
        \draw [->] (ep.south) -- ++(0,-0.5cm);

        \node [name=pmd,draw,rectangle,rounded corners] at (9, -4.5) {Photo-metric Distortion};
        \draw [->] (rc.south) -- (pmd.north);

        \node [name=mdl,draw,rectangle,rounded corners] at (9, -5.5) {Model};
        \draw [->] (pmd.south) -- (mdl.north);
    \end{tikzpicture}\end{center}
    \caption{Overview of StyleMixDG. Each source image is stylized offline using N randomly selected styles, yielding an N-times augmented dataset. During training, images are sampled with 80\% probability from the augmented set and 20\% from the original, followed by photometric distortion. }\label{fig:method-summary}
\end{figure}

\section{Introduction}
Deep learning models for computer vision often suffer from poor generalization when deployed in real-world settings, especially when trained on synthetic data due to the Sim2Real gap \cite{Zhao20Sim2Real}. 
This issue is particularly acute in autonomous driving, where collecting diverse, safety-critical datasets is costly or impractical. 
Critically, we operate under the domain generalization assumption, where no real-world target data is available during training. This distinguishes our setting from domain adaptation approaches and reflects scenarios where data collection is infeasible due to cost, safety, or privacy constraints.
To address this, one of the most widely used strategies is data augmentation, which aims to increase the diversity of the training set and improve robustness to distribution shifts.

Within this domain, style transfer has emerged as a promising technique~\cite{Peng21TextureRandomization,Lee22WildNet} to introduce controlled variation in the visual appearance of training samples. 
However, the literature contains unresolved contradictions that hamper the principled application of style transfer for domain generalization.
Prior studies differ in the size of the style pool employed (\eg, 8~in~\cite{Zheng19STaDA} or 15~in~\cite{Peng21TextureRandomization}), in whether texture complexity of the style images matters~\cite{Peng21TextureRandomization,Cakir24ST}, and in the choice of style source: artistic~\cite{Peng21TextureRandomization,Somavarapu20FrustratinglySimpleDG,Zheng19STaDA}, source-domain~\cite{Li23StyleInversion,Borlino21DGBaselines}, or real-world~\cite{Li23StyleInversion}. 
These discrepancies leave practitioners without clear guidance on how to configure style transfer augmentation.

The primary contribution of this work is a systematic empirical study that isolates and evaluates each of these design axes under controlled conditions, resolving conflicting findings in prior work. 
Concretely, we investigate: (1)~whether expanding the style pool leads to better generalization than repeated augmentation with a small set, (2)~whether
texture complexity of the style images affects domain robustness, and (3)~whether the type of style source has a measurable influence on performance.
Our study yields clear, actionable answers to all three questions.

Based on these empirical findings, we derive \emph{StyleMixDG} (Style-Mixing for Domain Generalization), a simple augmentation recipe that instantiates the design principles identified by our study. 
StyleMixDG is lightweight and model-agnostic: it does not require changes to the model architecture or training pipeline, nor does it introduce additional loss functions such as the alignment losses used in~\cite{Peng21TextureRandomization,Yue19PyramidConsistency,Lee22WildNet}. 
Instead, it increases data variability by mixing source-domain images with artistically stylized variants drawn from a large pool, augmenting each image multiple times, and applying online photometric distortions during training (see \cref{fig:method-summary}).
StyleMixDG thus serves as both a practical baseline and a validation that the empirically identified principles translate into competitive performance.

In summary, our contributions are: (1)~a systematic empirical analysis of style diversity, texture complexity, and style source in style transfer augmentation, resolving inconsistencies in prior work; (2)~actionable design principles for practitioners configuring style transfer for domain generalization; and (3)~StyleMixDG, a simple augmentation recipe derived from these principles that achieves competitive results on the GTAV $\to$ \{BDD100k, Cityscapes, Mapillary Vistas\} benchmark without architectural or training pipeline modifications.
\section{Related Work}\label{chap:preliminaries}

Style transfer, the process of applying the visual appearance of one image to another while preserving content \cite{Jing20NeuralStyleTransfer}, has been widely adopted for data augmentation \cite{Zheng19STaDA,Yang25StyleReplacement,Hong21StyleMix}. 
In the context of domain generalization (DG), several works employ style transfer to reduce overfitting to domain-specific statistics \cite{Yue19PyramidConsistency,Lee22WildNet,Somavarapu20FrustratinglySimpleDG,Peng21TextureRandomization}.
Yue et al. \cite{Yue19PyramidConsistency} and Lee et al. \cite{Lee22WildNet} use external real-world styles, while Peng et al. \cite{Peng21TextureRandomization} and Somavarapu et al. \cite{Somavarapu20FrustratinglySimpleDG} rely on artistic styles to maximize visual variance. 
Kim et al. \cite{Kim23WEDGE} explore weakly-labeled web data as both training input and style source, and Li et al. \cite{Li23StyleInversion} investigate source-domain and real-world texture styles. 
However, prior works use small style pools (\eg, 8 in \cite{Zheng19STaDA}, 15 in \cite{Peng21TextureRandomization}) and differ in their treatment of texture complexity and style source, leaving key design choices unresolved.
We build on this line of work by systematically analyzing these factors and proposing a lightweight augmentation framework that requires no architectural or loss modifications.

\section{Method}\label{chap:style-transfer}
We investigate the impact of diversity through style transfer in domain generalization. 
To this end, we build on a simple style transfer augmentation method presented in \cref{sec:gtr}.
We extend it by increasing diversity in multiple ways as described in \cref{sec:stylemixdg}

\subsection{Style Transfer Augmentation}\label{sec:gtr}
\begin{table}[tb]
    \caption{Examples of Style Transfer Augmentation (STA). This table shows examples of applying STA to synthetic images using style images of varying complexity. The first row depicts the style images taken from the Painter by Numbers\cite{painter-by-numbers} dataset. }\label{tab:gtr-demo}
  \begin{tabularx}{\linewidth}{XXXX}
    Content Image & Low Complexity & Medium Complexity & High Complexity\\\toprule
    & \includegraphics[width=2.5cm, height=2.5cm]{Resources/tcps/19492}
    & \includegraphics[width=2.5cm, height=2.5cm]{Resources/tcps/7129}
    & \includegraphics[width=2.5cm, height=2.5cm]{Resources/tcps/12271} \\
    
    
    \includegraphics[width=2.5cm, height=2.5cm]{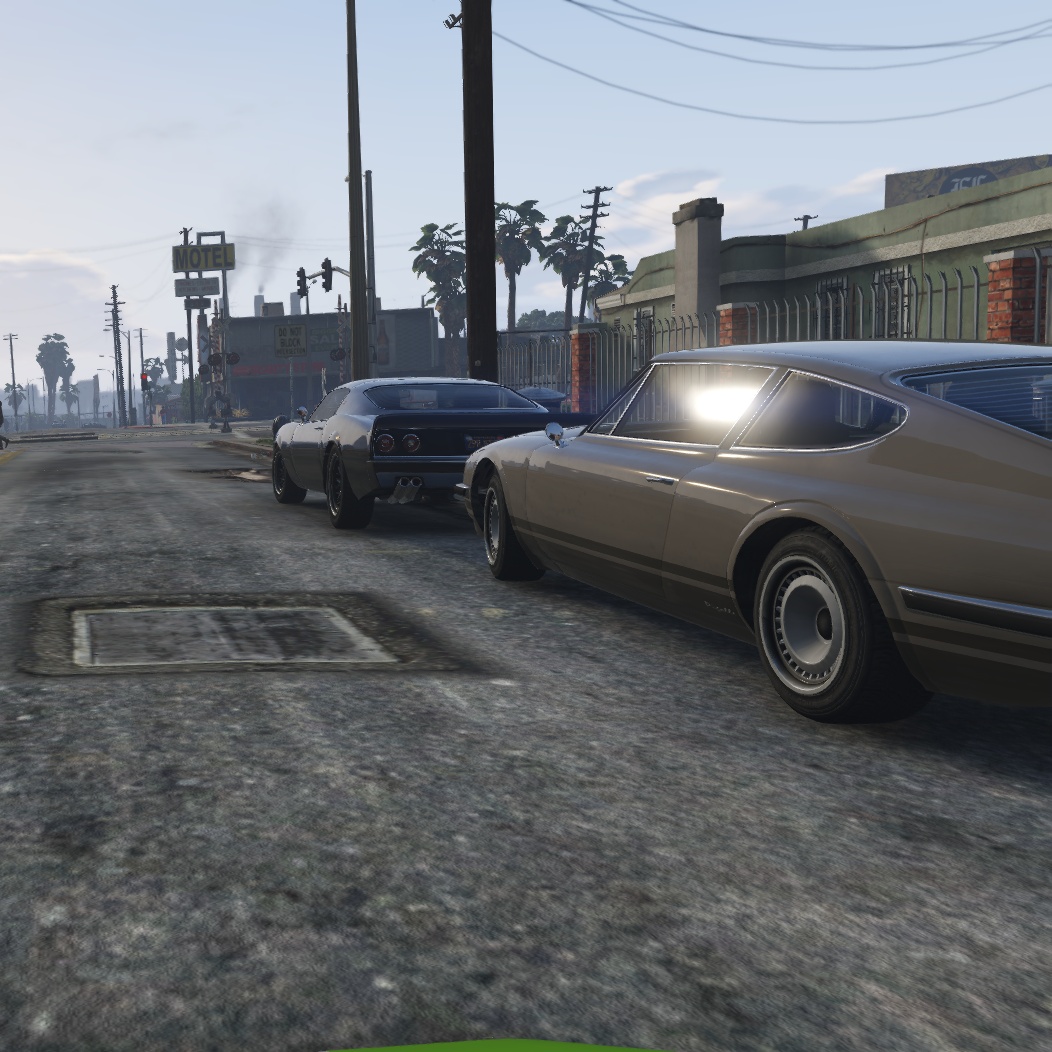}
    & \includegraphics[width=2.5cm, height=2.5cm]{Resources/gtr/20078_01_19492}
    & \includegraphics[width=2.5cm, height=2.5cm]{Resources/gtr/20078_01_7129}
    & \includegraphics[width=2.5cm, height=2.5cm]{Resources/gtr/20078_01_12271} \\
    
    \end{tabularx}

\end{table}

We employ style transfer as an offline data augmentation strategy, which we refer to as \textit{Style Transfer Augmentation} (STA), following \cite{Peng21TextureRandomization,Somavarapu20FrustratinglySimpleDG}. STA stylizes each input image using a randomly selected style image. This approach does not require any modifications to the model architecture or training procedure. Instead, models are trained on input images sampled from the style-augmented dataset.

We choose the Painter By Numbers dataset \cite{painter-by-numbers} as a source of artistic images.
This dataset comprises over 100.000 paintings from many artists and therefore provides a variability in style images.
Using artistic styles for randomization should introduce irregular and unrealistic textures into the training process.
As a consequence, the network should learn more domain-invariant cues such as shape and spatial layout, as it becomes harder to rely on texture.

As a method for style transfer, we use AdaIN \cite{Huang17AdaIn}, which performs style transfer by aligning channel-wise mean and variance of content features to those of the style image. The stylized features are then decoded back to pixel space.
Some examples of source, style, and stylized images are shown in \cref{tab:gtr-demo}.

\subsection{StyleMixDG}\label{sec:stylemixdg}
Building on the style transfer augmentation baseline described above, we introduce four extensions motivated by the empirical analysis in \cref{sec:ablations}, which we preview here for clarity. 
First, we sample from a substantially larger set of styles (10{,}000) than commonly used in prior work (\eg, 8~in~\cite{Zheng19STaDA} or 15~in~\cite{Peng21TextureRandomization}). 
Second, each image is augmented with multiple different style images rather than only one. 
Third, we apply online photometric distortion during training. 
Fourth, we mix style-augmented and original source-domain images at an 80:20 ratio.
The mixing ratio and number of stylized variants per image are hyperparameters selected in preliminary experiments. 
Each of these choices is empirically validated in \cref{sec:ablations}; their combination defines StyleMixDG and is summarized in \cref{fig:method-summary}.
\section{Experimental Setup}
\newcommand{\graycell}[1]{%
  \textcolor{gray}{#1}%
}
\begin{table}[t]
    \caption{State-of-the-Art Comparison. Shows the mIoU achieved using different networks on the three benchmark datasets (BDD, CS, MV) and their average. Source-only refers to standard training on synthetic data without any domain generalization technique applied. Methods above the gray divider rely solely on augmentation without modifying the model architecture or training pipeline; methods below require substantial architectural or loss-level changes. Results are either taken from~\cite{Peng22SemAwareDG} as denoted by~$^+$, from~\cite{Niemeijer23DIDEX} as denoted by~$^\to$, and otherwise from the original paper.}\label{tab:sota}
    \centering
        \begin{tabularx}{\textwidth}{lp{0.1cm}XXXXp{0.1cm}XXXX}
            Variant && \multicolumn{4}{c}{ResNet50} && \multicolumn{4}{c}{ResNet101}\\\cmidrule(lr){3-6}\cmidrule(lr){8-11}
             && BDD & CS & MV & Avg && BDD & CS & MV & Avg\\\toprule
            Source-only (Ours)                               && $17.20$ & $23.51$ & $33.62$ & $24.78$                                             && $24.83$ & $28.70$ & $37.92$ & $30.48$\\
            StyleMixDG (Ours)                           && $35.61$ & $40.75$ & $42.45$ & $39.54$         && $36.34$ & $42.74$ & $45.15$ & $41.41$\\
            \midrule
            \graycell{GTR \cite{Peng21TextureRandomization}}       && \graycell{$33.75^+$} & \graycell{$37.53^+$} & \graycell{$34.52^+$} & \graycell{$35.16^+$}                                     && \graycell{$39.60$} & \graycell{$43.70$} & \graycell{$39.10$} & \graycell{$40.80$}\\
            \graycell{SAN-SAW \cite{Peng22SemAwareDG}} && \graycell{$37.34$} & \graycell{$39.75$} & \graycell{$41.86$} & \graycell{$39.65$}                           && \graycell{$41.20^*$} & \graycell{$45.30^*$} & \graycell{$40.80^*$} & \graycell{$42.40^*$}\\
            \graycell{WildNet \cite{Lee22WildNet}}                  && \graycell{$38.42^+$} & \graycell{$44.62^+$} & \graycell{$46.09^+$} & \graycell{$43.04^+$} && \graycell{$41.70^*$} & \graycell{$45.80^*$} & \graycell{$47.10^*$} & \graycell{$44.90^*$}\\
            \graycell{DIDEX \cite{Niemeijer23DIDEX}}                &&&&&                                                                                && \graycell{$40.90^*$} & \graycell{$52.40^*$} & \graycell{$49.20^*$} & \graycell{$47.50^*$}\\\bottomrule
        \end{tabularx}

\end{table}

Having introduced our method and its key design elements, we investigate whether our lightweight style transfer method can be competitive to state-of-the-art domain generalization methods. 
In the following, we describe the benchmark and training settings used.

\subsection{Benchmark}\label{sec:benchmark}
The GTAV $\rightarrow$ \{BDD, CS, MV\} benchmark is a common proxy for assessment of domain generalization performance in semantic segmentation tasks\cite{Niemeijer23DIDEX,Peng22SemAwareDG,Yue19PyramidConsistency,Kim23WEDGE,Peng21TextureRandomization,Lee22WildNet}.
In this benchmark, neural networks are trained on the synthetic GTAV \cite{gtav} dataset for automotive semantic segmentation.
The datasets Berkeley Deep Drive 100k (BDD) \cite{bdd100k}, Cityscapes (CS) \cite{cityscapes} and Mapillary Vistas (MV) \cite{vistas} serve as real-world counterparts to GTAV, on which the networks will be evaluated.
Since these datasets' label spaces are not completely identical, they are mapped to a set of 19 common classes (see appendix \cref{tab:label-map-gtav}).

The benchmark performance is summarized using the average of the mIoU \cite{pascal-voc-2012} achieved by the same model on the three benchmark datasets.

We compare our method against a baseline of training on synthetic data without applying any domain generalization techniques, which we refer to as source-only training in the following.

\subsection{Training Settings}\label{sec:training-settings-gtav}
We employ ResNets \cite{He16ResNet} as backbones and DeepLabV2 \cite{deeplabv2} as semantic segmentation head as it is common for this benchmark \cite{Peng21TextureRandomization,Niemeijer23DIDEX}.

The backbones are initialized using ImageNet \cite{imagenet} weights.
We use stochastic gradient descent as optimizer with a batch size of 8, momentum of 0.9 and weight decay of $5e^{-4}$.
The learning rate is initially set to $5e^{-3}$ and scheduled using a polynomial learning rate scheduler \cite{Chen17AtrousConv} to $5e^{-5}$ at the end of training.
The networks are trained for $80.000$ iterations.
We use a weighted cross-entropy loss where class weights are computed per sample as one minus the pixel frequency of each class.

We pre-process the GTAV dataset by cropping each image into two patches of 1052x1052 pixels with minimal overlap and resizing them to 640x640 pixels.
In terms of augmentations, we employ random mirror and Gaussian blur with a probability of $50\%$.
For the latter, the radius is randomly chosen from the interval $(0, 1)$.

\section{Experiments}\label{sec:ablations}

In this section, we first benchmark our approach against state-of-the-art methods (\cref{sec:comparison-with-sota}), then conduct ablation experiments to validate our design choices and analyze how various aspects of style transfer influence domain generalization.
We start by comparing the effectiveness of the style transfer augmentation to standard augmentations in \cref{sec:exp-setup-results}. 
Following experiments investigate how increased variation contributes to generalization (\cref{sec:more_var}), and highlight the role of complexity and type of styles (\cref{sec:exp_texture_complexity} and \cref{sec:style-dataset}).
\Cref{sec:ablation-study} concludes the style transfer experiments by verifying the contribution of each component, followed by demonstrating transferability across architectures (\cref{sec:transformer}).

\begin{figure}[tb]
    \centering

    \begin{subfigure}{0.24\textwidth}\centering Image\end{subfigure}
    \begin{subfigure}{0.24\textwidth}\centering Groundtruth\end{subfigure}
    \begin{subfigure}{0.24\textwidth}\centering Source-only\end{subfigure}
    \begin{subfigure}{0.24\textwidth}\centering StyleMixDG\end{subfigure}

    \begin{subfigure}{0.24\textwidth}\includegraphics[width=\textwidth]{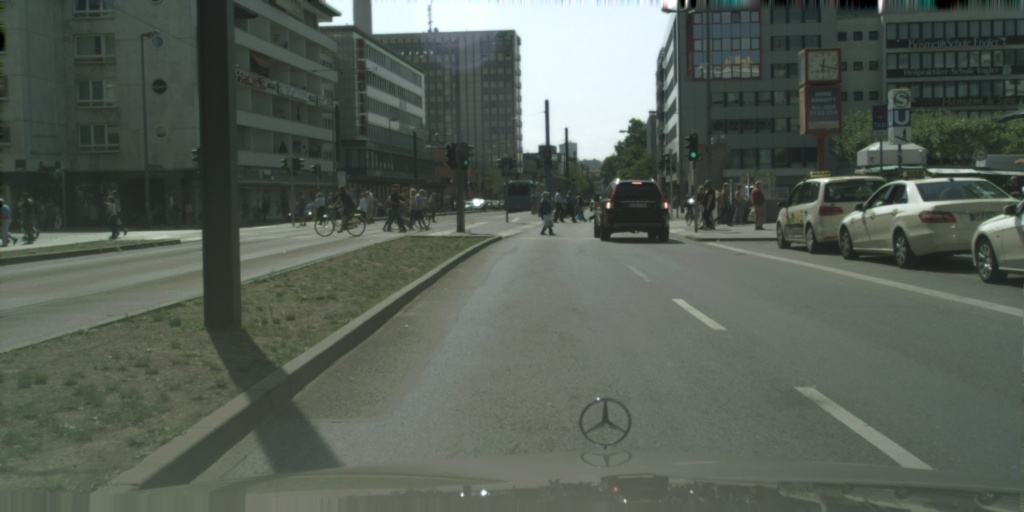}\end{subfigure}
    \begin{subfigure}{0.24\textwidth}\includegraphics[width=\textwidth]{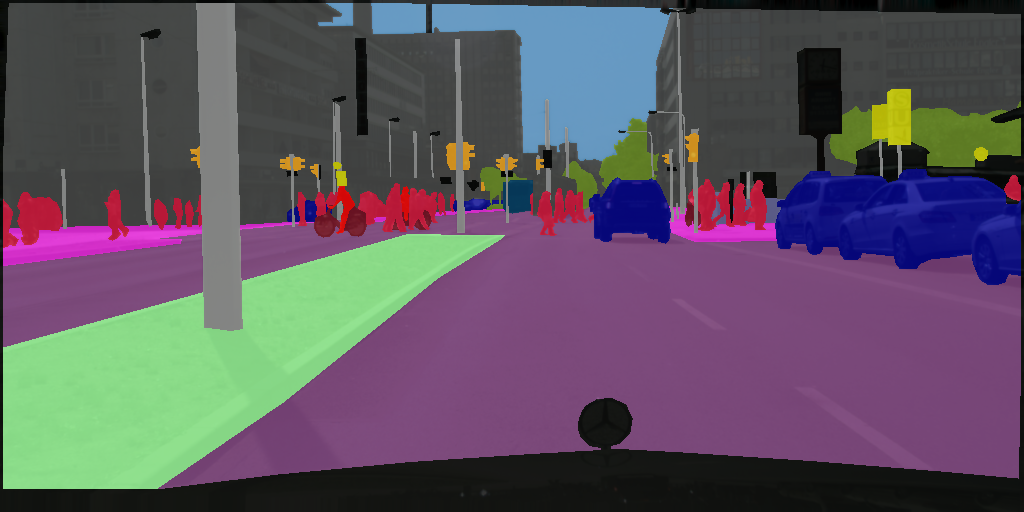}\end{subfigure}
    \begin{subfigure}{0.24\textwidth}\includegraphics[width=\textwidth]{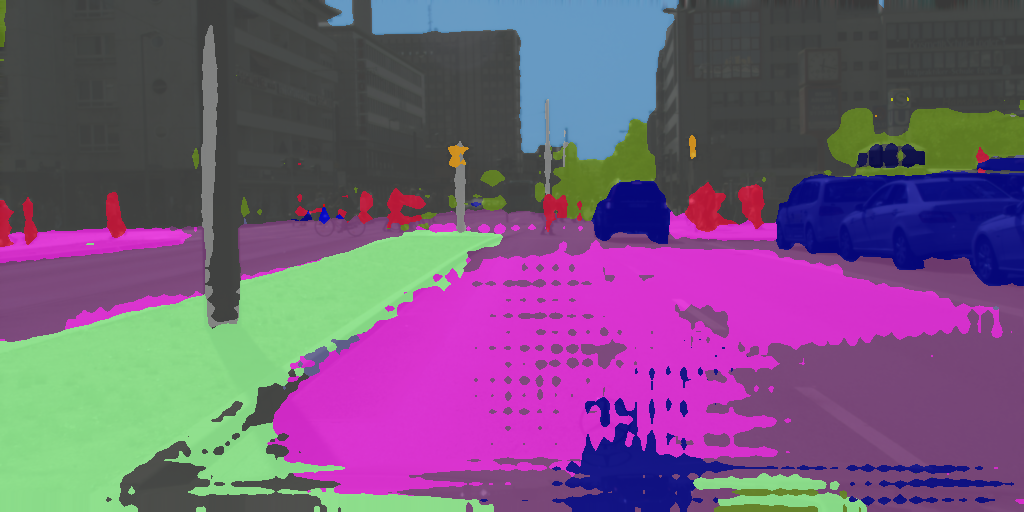}\end{subfigure}
    \begin{subfigure}{0.24\textwidth}\includegraphics[width=\textwidth]{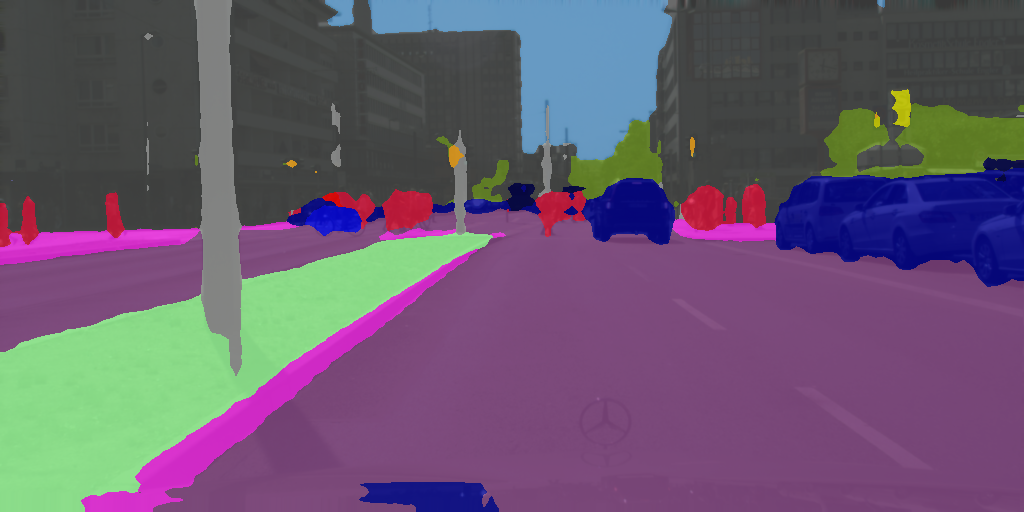}\end{subfigure}

    \begin{subfigure}{0.24\textwidth}\includegraphics[width=\textwidth]{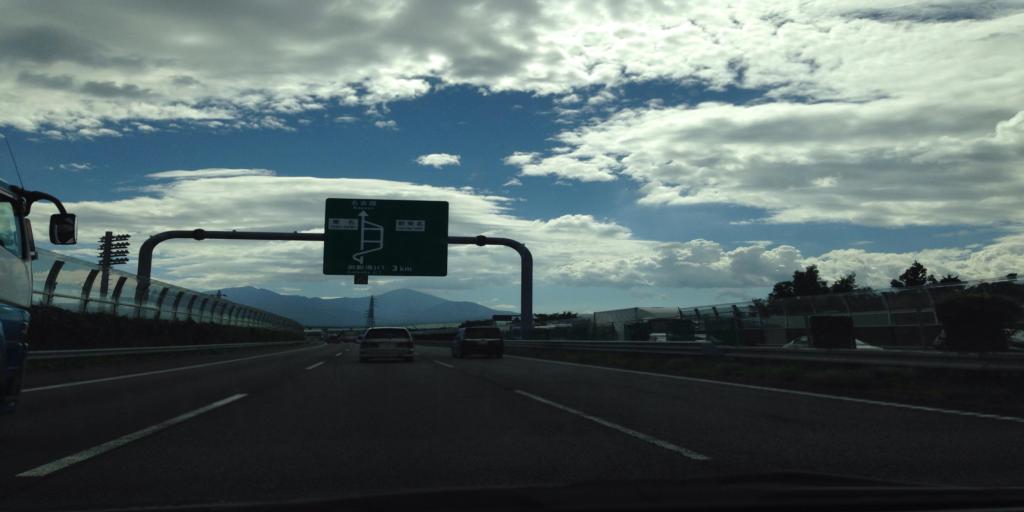}\end{subfigure}
    \begin{subfigure}{0.24\textwidth}\includegraphics[width=\textwidth]{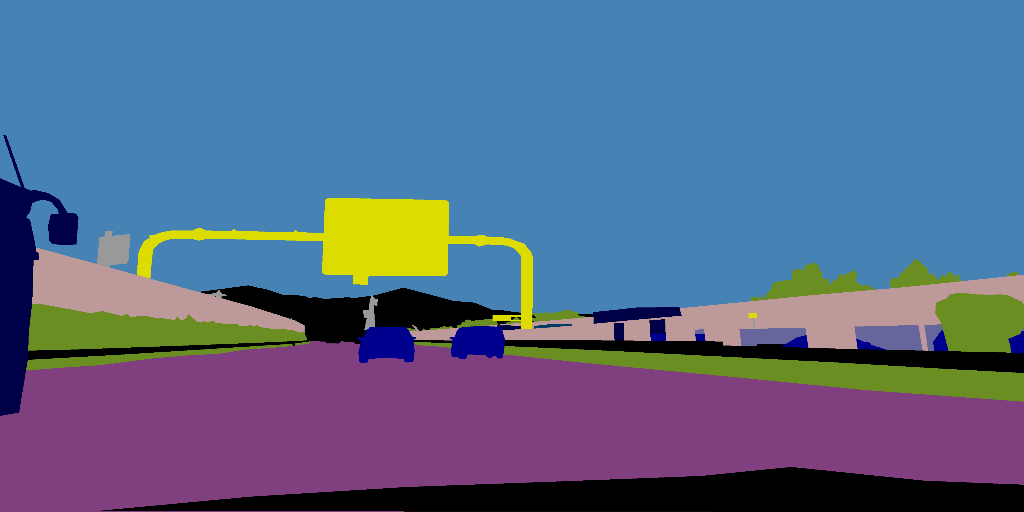}\end{subfigure}
    \begin{subfigure}{0.24\textwidth}\includegraphics[width=\textwidth]{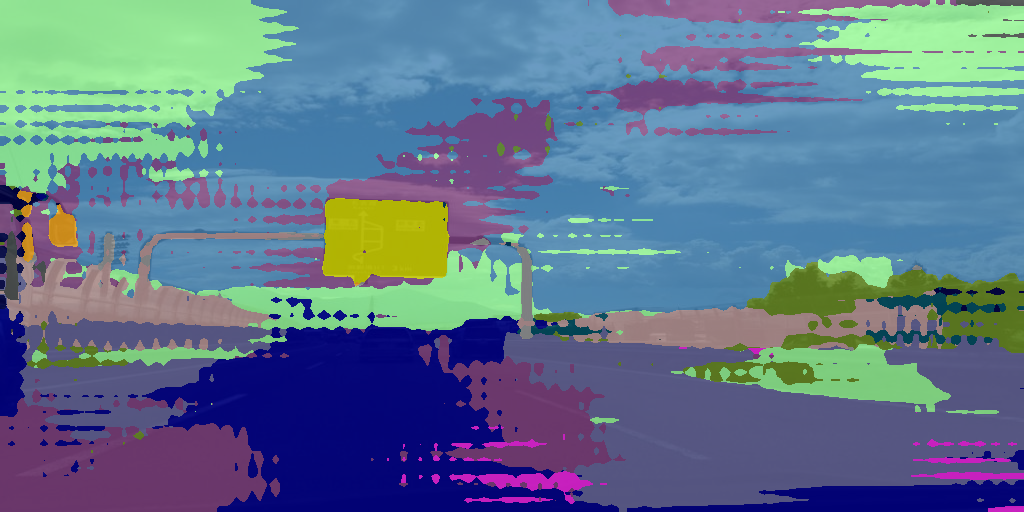}\end{subfigure}
    \begin{subfigure}{0.24\textwidth}\includegraphics[width=\textwidth]{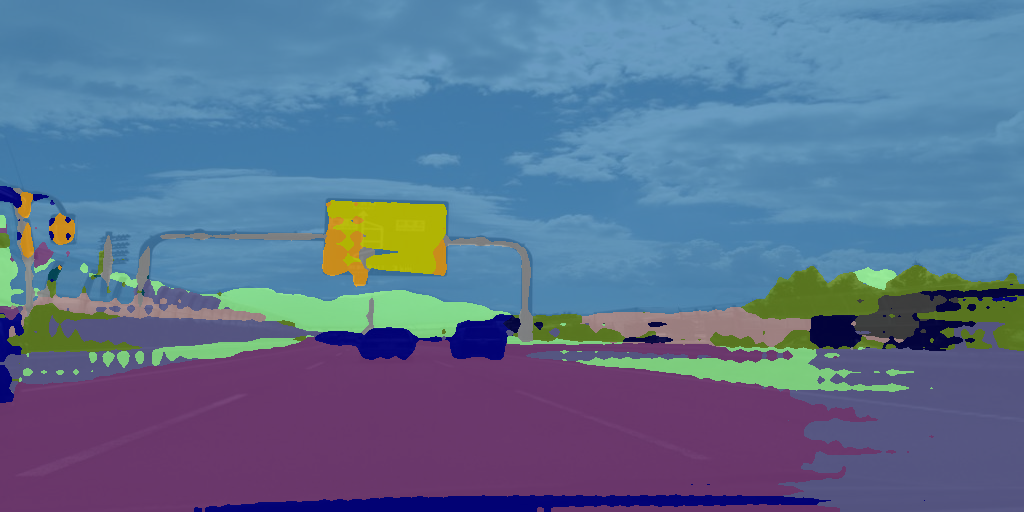}\end{subfigure}

    \begin{subfigure}{0.24\textwidth}\includegraphics[width=\textwidth]{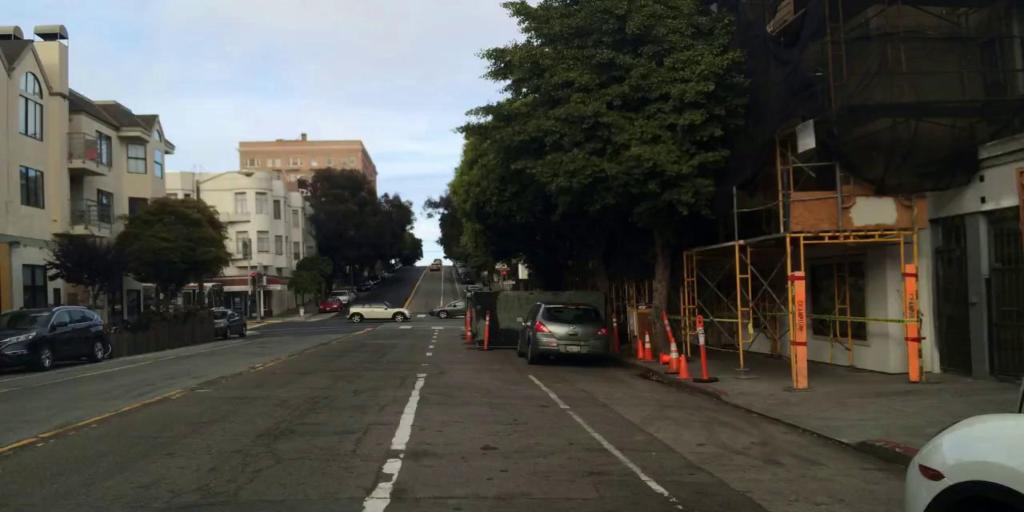}\end{subfigure}
    \begin{subfigure}{0.24\textwidth}\includegraphics[width=\textwidth]{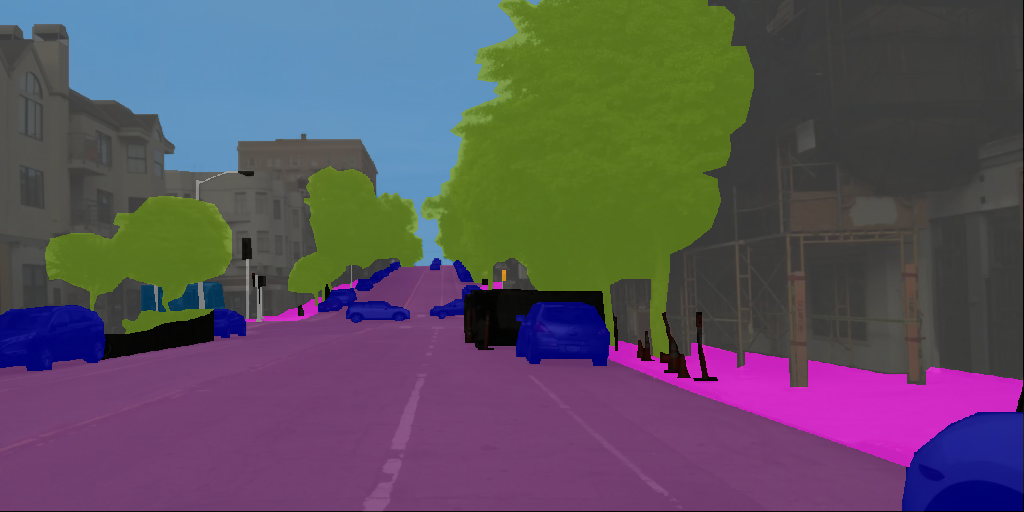}\end{subfigure}
    \begin{subfigure}{0.24\textwidth}\includegraphics[width=\textwidth]{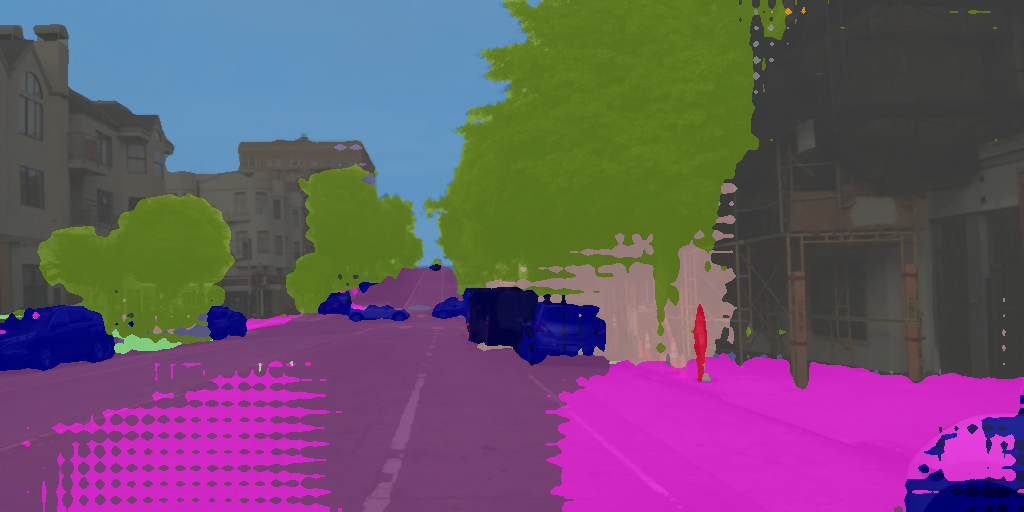}\end{subfigure}
    \begin{subfigure}{0.24\textwidth}\includegraphics[width=\textwidth]{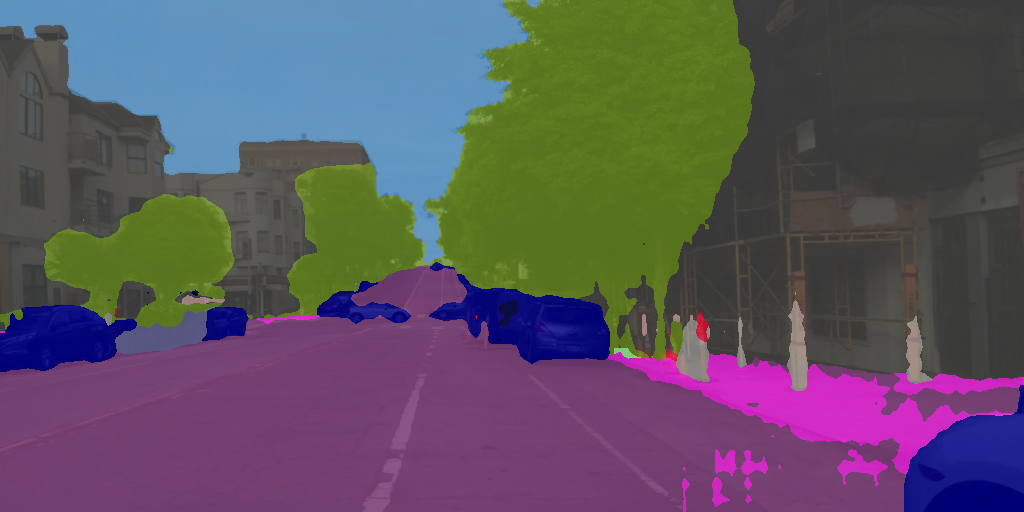}\end{subfigure}

    \begin{subfigure}{\textwidth}\centering\includegraphics[width=0.99\textwidth]{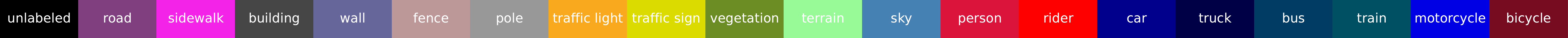}\end{subfigure}
    
    \caption{Qualitative Results. This figure shows qualitative results of applying the R101 Source-only and StyleMixDG variants to images from the target datasets.}\label{fig:sota-examples}
\end{figure}

\subsection{Results}\label{sec:comparison-with-sota}
Before presenting the detailed empirical analysis that motivates our design choices (\cref{sec:exp-setup-results} to \cref{sec:ablation-study}), we first show that the configuration they converge on, StyleMixDG, is competitive with the state of the art, establishing that the findings have practical relevance.

We compare StyleMixDG against state-of-the-art methods in \cref{tab:sota}\footnote{Per-class mIoUs for the ResNet-50 and ResNet-101 models can be found in appendix \cref{tab:detailed-results-r50} and \cref{tab:detailed-results-r101} respectively.}.
With ResNet-50, StyleMixDG achieves an average mIoU of $39.54$, significantly outperforming GTR \cite{Peng21TextureRandomization} ($35.16$), which is the method most comparable to ours. 
This is notable because GTR requires training on three differently augmented images simultaneously with a consistency loss, substantially increasing computational cost, whereas StyleMixDG relies solely on augmentation through simple random sampling. 
With ResNet-101, StyleMixDG remains competitive ($41.41$ vs. GTR's $40.80$ average mIoU), though the per-dataset picture is mixed: StyleMixDG leads on Cityscapes and Mapillary Vistas, while GTR achieves a higher score on BDD100K. The narrowing gap suggests that gains from data-driven augmentation alone may saturate with increasing model capacity. 
Methods such as SAN-SAW \cite{Peng22SemAwareDG}, WildNet \cite{Lee22WildNet} and DIDEX \cite{Niemeijer23DIDEX} achieve higher scores but rely on substantially more complex mechanisms including alignment modules, additional losses, and modified training pipelines. 
In contrast, StyleMixDG is entirely model-agnostic, easy to integrate and comes at low computational cost. 
Qualitative results are shown in \cref{fig:sota-examples}.

In the following sections, we iteratively build our method, starting with the style transfer augmentation baseline.

\subsection{Baseline}\label{sec:exp-setup-results}

To ground subsequent improvements, we compare three ResNet-18 training variants: source-only training, Style Transfer Augmentation (STA) with 15 styles following \cite{Peng21TextureRandomization}, and Photometric Distortion (PMD) \cite{mmseg}.
PMD is a standard color-jitter augmentation applying random brightness, saturation, hue and contrast changes. 
Unless otherwise noted, all following sections use ResNet-18 and report mean $\pm$ standard deviation over three runs with different seeds and, for style transfer variants, differently augmented datasets.
As shown in \cref{tab:st-variation}, STA improves over source-only training, consistent with prior work \cite{Peng21TextureRandomization}. 
However, PMD performs competitively despite its simplicity, raising the question of whether the added cost of offline stylization can be justified. 
The following sections show how increasing stylistic diversity addresses this.

\subsection{More Variation, More Robustness} \label{sec:more_var}

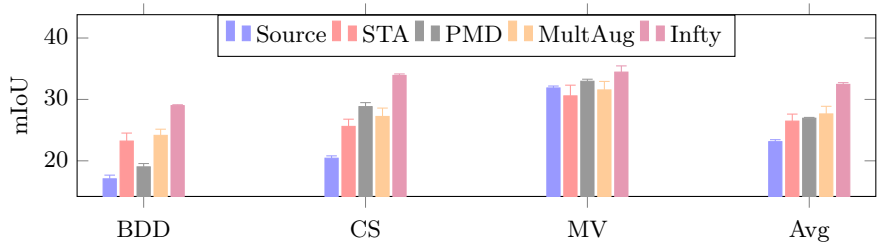
\begin{figure}[tb]   

    \begin{tikzpicture}
     
    \begin{axis} [
        width = \textwidth,
        height = 4cm,
        ylabel={mIoU},
        ybar = .05cm,
        bar width = 5pt,
        ymin = 15,
        ymax = 43,
        ytick = {20, 30, 40, 50, 60},
        symbolic x coords = {BDD, CS, MV, Avg},
        xtick = data,
        xticklabel style={name=tick no \ticknum},
        typeset ticklabels with strut,
        enlarge x limits = 0.1,
        enlarge y limits = {abs = .8},
        legend style={
            at={(0.5,1.0)},         
            anchor=north,            
            cells={align=left}
        },
        legend columns=6,
    ]

    \addplot+ [blue!40!white,error bars/.cd, y dir=plus, y explicit, ] coordinates {
        (BDD,17.07) +- (0,0.6)
        (CS,20.43) +- (0,0.38)
        (MV,31.87) +- (0,0.31)
        (Avg,23.12) +- (0,0.32)
    };

    \addplot+ [red!40!white, error bars/.cd, y dir=plus, y explicit, ]  coordinates {
        (BDD,23.22)  +- (0,1.3)
        (CS,25.61) +- (0,1.17)
        (MV,30.58) +- (0,1.74)
        (Avg,26.47) +- (0,1.14)
    };


    \addplot+ [black!40!white, error bars/.cd, y dir=plus, y explicit]  coordinates {
        (BDD,19.02) +- (0,0.54)
        (CS,28.83) +- (0,0.65)
        (MV,32.96) +- (0,0.33)
        (Avg,26.94) +- (0,0.13)
    };

    \addplot+ [orange!40!white, error bars/.cd, y dir=plus, y explicit]  coordinates {
        (BDD,24.15) +- (0,1.0)
        (CS,27.23) +- (0,1.36)
        (MV,31.55) +- (0,1.38)
        (Avg,27.64) +- (0,1.24)
    };

    \addplot+ [purple!40!white, error bars/.cd, y dir=plus, y explicit]  coordinates {
        (BDD,28.99) +- (0,0.14)
        (CS,33.91) +- (0,0.24)
        (MV,34.45) +- (0,1.01)
        (Avg,32.45) +- (0,0.28)
    };
     
    \legend {
        Source, 
        STA, 
        PMD, 
        MultAug, 
        Infty
    };
     
    \end{axis}
     
    \end{tikzpicture}
    
    \caption{Results using more variation in style transfer. Source: training without style-transfer augmentation. STA is the baseline Style Transfer Augmentation. It is compared against Photo-metric Distortion (PMD), multiple style augmentations per image (MultAug) and sampling style images from a large pool of styles (Infty). Shows the mIoU achieved using different networks on the three benchmark datasets (BDD, CS, MV) and their average.}\label{tab:st-variation}
\end{figure}

Standard STA is limited by applying stylization only once per image and drawing from a small style pool. 
We explore two extensions: (i) STA-MultAug, which generates three stylized versions per image from the same 15 styles and randomly samples one per iteration, and (ii) STA-Infty, which samples from 10,000 styles while keeping one stylized version per image. 
As shown in \cref{tab:st-variation}, both variants outperform STA and PMD.
STA-Infty yields the largest improvement, confirming that style pool diversity has a stronger effect than repeated augmentation of the same styles. 
Crucially, the clear advantage of STA-Infty over PMD demonstrates that style transfer provides benefits beyond what standard color augmentations can achieve, justifying its additional pre-processing cost.

\subsection{Texture Complexity is Irrelevant} \label{sec:exp_texture_complexity}

Style images of varying texture complexity produce qualitatively different stylizations (\cf \cref{tab:gtr-demo}): low-complexity styles create washed-out textures, while high-complexity styles may introduce artifacts.
To test whether complexity affects generalization, we apply the Texture Complexity-based Painting Selection (TCPS) method from \cite{Peng21TextureRandomization}, which scores images by their ratio of small gradient pixels to total pixels. 
We partition the style dataset into low ([0, 0.5)), medium ([0.5, 0.75)) and high ([0.75, 1.0]) complexity subsets, sample 10,000 images from each, and train with STA. 
As shown in \cref{fig:st-complexity}, all three variants achieve nearly identical performance ($32.10$, $32.17$, $32.21$ mIoU), indicating that texture complexity has no significant effect when sampling from a sufficiently large pool.
This contrasts with the suggestion in \cite{Peng21TextureRandomization} that complexity matters.
We attribute the discrepancy to pool scale: at 10,000 styles, individual complexity effects average out, making filtering unnecessary.
We therefore exclude complexity-based selection from subsequent experiments.

\begin{table}[tb]
    \caption{Results using different texture complexities. Compares 
    sampling style images from low, medium and high-complexity styles 
    with no complexity-based selection (Infty). The number of styles 
    in the pool is identical. Shows the mIoU achieved using different 
    networks on the three benchmark datasets (BDD, CS, MV) and their 
    average.}\label{fig:st-complexity}
    \centering
    \begin{tabular}{lcccc}
        \hline
        Variant & BDD & CS & MV & Avg \\
        \hline
        Infty  & $28.99 \pm 0.14$ & $33.91 \pm 0.24$ & $34.45 \pm 1.01$ & $32.45 \pm 0.28$ \\
        Low    & $28.11 \pm 0.70$ & $34.35 \pm 0.25$ & $33.83 \pm 0.28$ & $32.10 \pm 0.17$ \\
        Medium & $28.46 \pm 0.97$ & $33.25 \pm 0.83$ & $34.81 \pm 1.02$ & $32.17 \pm 0.76$ \\
        High   & $29.22 \pm 1.24$ & $33.58 \pm 1.64$ & $33.84 \pm 0.89$ & $32.21 \pm 0.94$ \\
        \hline
    \end{tabular}
\end{table}

\subsection{The Choice of Style Dataset Matters}\label{sec:style-dataset}

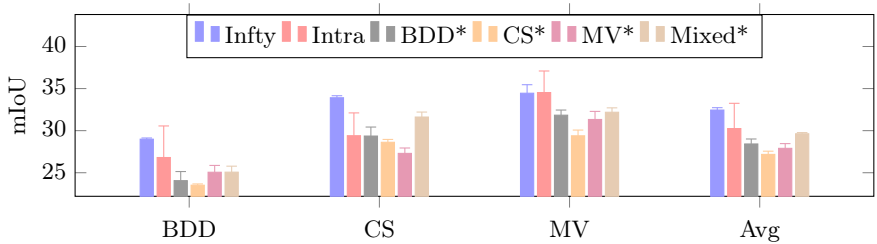
\begin{figure}[tb]
    \begin{tikzpicture}
     
    \begin{axis} [
        width = \textwidth,
        height = 4cm,
        ylabel={mIoU},
        ybar = .05cm,
        bar width = 5pt,
        ymin = 23,
        ymax = 43,
        ytick = {25, 30, 35, 40},
        symbolic x coords = {BDD, CS, MV, Avg},
        xtick = data,
        xticklabel style={name=tick no \ticknum},
        typeset ticklabels with strut,
        enlarge x limits = 0.2,
        enlarge y limits = {abs = .8},
        legend style={
            at={(0.5,1.0)},         
            anchor=north,            
            cells={align=left}
        },
        legend columns=6,
    ]


    \addplot+ [blue!40!white, error bars/.cd, y dir=plus, y explicit]  coordinates {
        (BDD,28.99) +- (0,0.14)
        (CS,33.91) +- (0,0.24)
        (MV,34.45) +- (0,1.01)
        (Avg,32.45) +- (0,0.28)
    };

    \addplot+ [red!40!white, error bars/.cd, y dir=plus, y explicit]  coordinates {
        (BDD,26.81) +- (0,3.75)
        (CS,29.40) +- (0,2.71)
        (MV,34.53) +- (0,2.56)
        (Avg,30.25) +- (0,3.00)
    };

    \addplot+ [black!40!white, error bars/.cd, y dir=plus, y explicit]  coordinates {
        (BDD,24.06) +- (0,1.08)
        (CS,29.36) +- (0,1.07)
        (MV,31.84) +- (0,0.61)
        (Avg,28.42) +- (0,0.59)
    };

    \addplot+ [orange!40!white, error bars/.cd, y dir=plus, y explicit]  coordinates {
        (BDD,23.52) +- (0,0.15)
        (CS,28.62) +- (0,0.34)
        (MV,29.39) +- (0,0.68)
        (Avg,27.18) +- (0,0.38)
    };

    \addplot+ [purple!40!white, error bars/.cd, y dir=plus, y explicit]  coordinates {
        (BDD,25.06) +- (0,0.81)
        (CS,27.30) +- (0,0.64)
        (MV,31.33) +- (0,0.96)
        (Avg,27.90) +- (0,0.56)
    };

    \addplot+ [brown!40!white, error bars/.cd, y dir=plus, y explicit]  coordinates {
        (BDD,25.07) +- (0,0.71)
        (CS,31.62) +- (0,0.59)
        (MV,32.19) +- (0,0.52)
        (Avg,29.63) +- (0,0.14)
    };
     
    \legend {
        Infty, 
        Intra,
        BDD*,
        CS*,
        MV*,
        Mixed*
    };
     
    \end{axis}
     
    \end{tikzpicture}
    \caption{Results using different style sources. Style sources include the Painter-by-Numbers (Infty), GTAV (Intra) and benchmark datasets (BDD, CS, MV). Mixed denotes that all benchmark datasets are used to sample styles. Shows the mIoU achieved using different networks on the three benchmark datasets (BDD, CS, MV) and their average. The unsupervised domain adaptation settings are marked with an asterisk.}\label{tab:st-texture}
\end{figure}

While style transfer is widely used for domain generalization, prior works differ in their choice of style source: some rely on artistic styles \cite{Peng21TextureRandomization,Somavarapu20FrustratinglySimpleDG,Zheng19STaDA}, others on source-domain styles \cite{Li23StyleInversion,Borlino21DGBaselines} or real-world textures \cite{Li23StyleInversion}, yet the impact of this choice has not been systematically compared. 
We address this by evaluating artistic styles (STA-Infty) against source-domain stylization (STA-Intra, sampling styles from GTAV) and target-domain stylization (using BDD, CS, MV individually and combined). 
The latter constitutes unsupervised domain adaptation (UDA), relaxing the DG setting for this analysis. 
As shown in \cref{tab:st-texture} STA-Infty outperforms all domain-aligned variants. 
STA-Intra performs competitively but with higher variance. 
This suggests that a broad range of artistic styles, which introduce maximal distributional variation, is more effective than domain-aligned styles for generalization.

\subsection{Ablation Study}\label{sec:ablation-study}

Having shown detailed investigations on the influence of style diversity, complexity and source, we provide an ablation of our method StyleMixDG.
\Cref{tab:st-combinations} shows that all building blocks contribute small improvements. 
Their combination leads to the strongest generalization performance observed in our experiments.

\begin{table}[tb]
    \caption{Results using combinations of previous methods. Infty: increase the number of styles to sample from, Mixing: mix non-stylized and stylized images, MultAug: use multiple styles per image, PMD: apply photo-metric distortion. Shows the mIoU achieved using different networks on the three benchmark datasets (BDD, CS, MV) and their average. The best and second-best entries are highlighted in \textbf{bold} and \textit{italics} respectively. }\label{tab:st-combinations}
  \centering
    \begin{tabularx}{\textwidth}{ccccXXXX}
        Infty & Mixing & MultAug & PMD & BDD & CS & MV & Average\\\toprule
        \xmark&\xmark&\xmark&\xmark& $17.07\pm 0.60$ & $20.43\pm 0.38$ & $31.87\pm 0.31$ & $23.12\pm 0.32$\\
        \cmark&\xmark&\xmark&\xmark& $28.99\pm 0.14$ & $33.91\pm 0.24$ & $34.45\pm 1.01$ & $32.45\pm 0.28$\\
        \cmark&\xmark&\xmark&\cmark& $29.22\pm 0.86$ & $34.87\pm 0.89$ & $34.58\pm 0.45$ & $32.89\pm 0.35$\\
        \cmark&\cmark&\xmark&\xmark& $29.13\pm 0.84$ & $33.57\pm 1.02$ & $\mathbf{36.19\pm 0.66}$ & $32.97\pm 0.47$\\
        \cmark&\xmark&\cmark&\xmark& $29.09\pm 0.50$ & $34.91\pm 0.13$ & $34.77\pm 0.24$ & $32.92\pm 0.20$\\
        \cmark&\cmark&\cmark&\xmark& $\mathit{29.81\pm 0.12}$ & $\mathit{36.03\pm 0.39}$ & $35.66\pm 0.65$ & $\mathit{33.83\pm 0.30}$\\
        \cmark&\cmark&\cmark&\cmark& $\mathbf{30.28\pm 0.68}$ & $\mathbf{37.12\pm 0.23}$ & $\mathit{35.85\pm 0.76}$ & $\mathbf{34.41\pm 0.45}$\\
        \bottomrule
    \end{tabularx}
     
\end{table}

\subsection{Extension to Transformers}\label{sec:transformer}

To verify that our findings transfer beyond CNNs, we train an UperNet \cite{upernet} segmentation head with both ResNet-50 and DeiT-S16 \cite{deit} backbones, following the training configuration of MMSegmentation \cite{mmseg}. 
DeiT-S16 is a vision transformer of comparable size to ResNet-50 ($22.05$M vs. $25.56$M parameters). 
As shown in \cref{tab:transformer}, StyleMixDG improves the DeiT-S16 baseline from $35.14$ to $40.00$ mIoU, confirming its effectiveness on transformer architectures. 
Interestingly, PMD slightly degrades DeiT-S16 performance on BDD100K and Mapillary Vistas while improving it on Cityscapes. 
This is consistent with findings that vision transformers exhibit reduced texture bias compared to CNNs \cite{Angarano24BackToBones}, making color-targeting augmentations less beneficial or even counterproductive. StyleMixDG, which introduces broader distributional variation beyond color statistics, remains effective across both architecture families.

\begin{table}[tb]
    \caption{Architecture Comparison. Source: non-stylized dataset, PMD: photo-metric distortion is applied, StyleMixDG: our method. Shows the mIoU achieved using different networks on the three benchmark datasets (BDD, CS, MV) and their average. The best entries are highlighted in \textbf{bold}. }\label{tab:transformer}
    \centering
        \begin{tabularx}{\textwidth}{lp{0.1cm}XXXXp{0.1cm}XXXX}
            Variant && \multicolumn{4}{c}{ResNet50 UperNet} && \multicolumn{4}{c}{DeIT-S16 UperNet}\\\cmidrule(lr){3-6}\cmidrule(lr){8-11}
             && BDD & CS & MV & Avg && BDD & CS & MV & Avg\\\toprule
            Source        && $30.54$ & $28.79$ & $35.04$ & $31.46$ && $33.74$ & $33.04$ & $38.63$ & $35.14$\\
            PMD           && $34.56$ & $39.83$ & $39.30$ & $37.89$ && $32.87$ & $\mathbf{38.69}$ & $37.80$ & $36.45$\\
            StyleMixDG    && $\mathbf{38.23}$ & $\mathbf{41.07}$ & $\mathbf{42.75}$ & $\mathbf{40.68}$         && $\mathbf{39.52}$ & $38.19$ & $\mathbf{42.29}$ & $\mathbf{40.00}$\\\bottomrule
        \end{tabularx}
    
\end{table}

\section{Limitations}

Style transfer can degrade small objects or render texture-distinguished classes indistinguishable, and rare classes such as \emph{train} show limited improvement under StyleMixDG (\cf \cref{tab:detailed-results-r50} and \cref{tab:detailed-results-r101}). 
Our evaluation is restricted to a single source domain (GTAV) and to semantic segmentation. 
Extending to additional synthetic sources, other perception tasks such as object detection, and mixed synthetic-real training regimes are natural directions for future work.

\section{Conclusion}
We presented a systematic empirical study of style transfer augmentation for domain generalization in autonomous driving, addressing three open questions in the literature. 
Our controlled experiments show that (i)~expanding the style pool has a larger effect than repeated augmentation with few styles, (ii)~texture complexity filtering is unnecessary when the pool is sufficiently large, and (iii)~diverse artistic styles outperform domain-aligned alternatives.
These findings resolve conflicting claims in prior work and provide actionable guidance for practitioners applying style transfer to domain generalization. 
As a practical instantiation of these principles, we derived StyleMixDG, a lightweight, model-agnostic augmentation recipe that achieves competitive results on the GTAV $\to$ \{BDD100k, Cityscapes, Mapillary Vistas\} benchmark without requiring architectural or training pipeline modifications.
This confirms that principled configuration of a simple augmentation strategy can rival more complex domain generalization methods.

\bibliographystyle{splncs04}
\bibliography{main}

\clearpage
\appendix
\section{Implementation Details}
\begin{table}
\caption{Label Correspondences. Label correspondences used in our experiments. The \emph{Label} column additionally shows the color for each class used in subsequent visualization.}\label{tab:label-map-gtav}
\begin{tabularx}{\textwidth}{llXX}
    GTAV\cite{gtav} & BDD\cite{bdd100k} & CS\cite{cityscapes} & MV\cite{vistas}\\\toprule
    \begin{filecontents*}{Resources/labels_gtav.csv}
class; color; bdd100k; cityscapes; vistas; gtav; idd
road; 128, 64, 128; road; road; bike lane, crosswalk - plain, lane marking - crosswalk, lane marking - general, manhole, parking, road, service lane; road; road
sidewalk; 244, 35, 232; sidewalk; sidewalk; curb, curb cut, pedestrian area, sidewalk; sidewalk; curb, sidewalk
building; 70, 70, 70; building; building; building; building; building
wall; 102, 102, 156; wall; wall; wall; wall; wall
fence; 190, 153, 153; fence; fence; fence; fence; fence
pole; 153, 153, 153; pole; pole, polegroup; pole, street light, utility pole; pole; pole, pole group
traffic light; 250, 170, 30; traffic light; traffic light; traffic light; traffic light; traffic light
traffic sign; 220, 220, 0; traffic sign; traffic sign; traffic sign (back), traffic sign (front), traffic sign frame; traffic sign; traffic sign
vegetation; 107, 142, 35; vegetation; vegetation; vegetation; vegetation; vegetation
terrain; 152, 251, 152; terrain; terrain; terrain; terrain; non-drivable fallback
sky; 70, 130, 180; sky; sky; sky; sky; sky
person; 220, 20, 60; pedestrian; person; person; person; person
rider; 255, 0, 0; rider; rider; bicyclist, motorcyclist, other rider; rider; rider
car; 0, 0, 142; car; car; car; car; autorickshaw, car
truck; 0, 0, 70; truck; truck; truck; truck; truck
bus; 0, 60, 100; bus; bus; bus; bus; bus
train; 0, 80, 100; train; train; on rails; train; train
motorcycle; 0, 0, 230; motorcycle; motorcycle; motorcycle; motorcycle; motorcycle
bicycle; 119, 11, 32; bicycle; bicycle; bicycle; bicycle; bicycle
unlabeled; 0, 0, 0; unlabeled; bridge, caravan, dynamic, ego vehicle, ground, guard rail, license plate, out of roi, parking, rail track, rectification border, static, trailer, tunnel, unlabeled; banner, barrier, bench, bike rack, billboard, bird, boat, bridge, car mount, caravan, catch basin, cctv camera, ego vehicle, fire hydrant, ground animal, guard rail, junction box, mailbox, mountain, other vehicle, phone booth, pothole, rail track, sand, snow, trailer, trash can, tunnel, unlabeled, water, wheeled slow; unlabeled; animal, billboard, bridge, caravan, drivable fallback, ego vehicle, fallback background, guard rail, license plate, obs-str-bar-fallback, out of roi, parking, rail track, rectification border, trailer, tunnel, unlabeled, vehicle fallback
\end{filecontents*}
\csvreader[
        separator=semicolon,
        range=1-15,
        late after line=\\\midrule,
        late after last line=\\\midrule
    ]{Resources/labels_gtav.csv}{}
    {\color[RGB]{\csvcolii}\rule{.5cm}{.5cm}~\normalcolor \csvcolvi & \csvcoliii & \csvcoliv & \csvcolv}
    \multicolumn{4}{l}{\textit{Table continues on the next page.}}\\\bottomrule
\end{tabularx}
\end{table}
\begin{table}
\ContinuedFloat
\caption{Label Correspondences. Label correspondences used in our experiments. The \emph{Label} column additionally shows the color for each class used in subsequent visualization.}\label{tab:label-map-gtav}
\begin{tabularx}{\textwidth}{llXX}
    GTAV\cite{gtav} & BDD\cite{bdd100k} & CS\cite{cityscapes} & MV\cite{vistas}\\\toprule
    \csvreader[
        separator=semicolon,
        range=16-20,
        late after line=\\\midrule,
        late after last line=\\\bottomrule
    ]{Resources/labels_gtav.csv}{}
    {\color[RGB]{\csvcolii}\rule{.5cm}{.5cm}~\normalcolor \csvcolvi & \csvcoliii & \csvcoliv & \csvcolv}
\end{tabularx}
\end{table}

\subsection{Style Transfer Augmentation}
We use the style transfer implementation and pre-trained networks from \url{https://github.com/naoto0804/pytorch-AdaIN}.
We only select paintings that have a width and height of $512$ pixels or more.
Before encoding the style images, we take the largest square-center crop of the style image and resize it to $512x512$ pixels.
We extract the features of the content images at the original size, \ie $1052x1052$ pixels.
As the decoded image has a resolution of $1056x1056$ pixels, we resize it to match the original size.

\subsection{Texture Complexity-based Painting Selection}
Pixels are classified as smooth if their gradient representation is smaller than a threshold $\epsilon=20$.
As the resolution of the paintings influences the area and therefore also the complexity, we use uniform painting resolutions for complexity computation.
In particular, we use the largest square-center crop and resize it to a resolution of $512x512$ pixels.
After resizing, we convert the images to grayscale.
We calculate the first discrete difference along the spatial axes to compute the pixel gradients:
$$\textnormal{grad}_x\left(i, j\right) = I\left(i+1, j\right) - I\left(i, j\right)$$
$$\textnormal{grad}_y\left(i,j\right) = I\left(i, j+1\right) - I\left(i, j\right)$$
We apply zero padding to the size of $512$ pixels squared.
The gradients in the horizontal and vertical direction are combined by summing the squared gradients:
$$\textnormal{grad}\left(i, j\right) = \textnormal{grad}^2_x\left(i, j\right) + \textnormal{grad}^2_y\left(i, j\right).$$
Using this procedure, we reproduce the smooth versus unsmooth area separation shown by Peng et al. \cite{Peng21TextureRandomization} in \cref{tab:smooth-vs-unsmooth-area}.
It reproduces the smooth versus unsmooth area separation originally shown in GTR\cite{Peng21TextureRandomization} to validate that our TCPS implementation yields consistent results. Since GTR does not provide sufficient implementation details, this replication serves as a verification step for our own implementation.

\begin{table*}
\begin{tabularx}{\textwidth}{lXXX}

    & Original & Smooth Area & Unsmooth Area\\\toprule
    \begin{filecontents*}{Resources/tcps.csv}
Image, Complexity
1903, 0.6953
19492, 0.1355
7129, 0.6098
12271, 0.8124
14405, 0.1537
13473, 0.1411
1425, 0.5252
12582, 0.6923
13170, 0.7896
11151, 0.8427
\end{filecontents*}
\csvreader[
        separator=comma,
        range=5-10,
        late after line=\\\midrule,
        late after last line=\\\bottomrule
    ]{Resources/tcps.csv}{}
    {
        \rotatebox{90}{\shortstack{\csvcoli \\ (\csvcolii)}} &
        \includegraphics[width=3cm, height=2.5cm]{Resources/tcps/\csvcoli} &
        \includegraphics[width=3cm, height=2.5cm]{Resources/tcps/\csvcoli_smooth} &
        \includegraphics[width=3cm, height=2.5cm]{Resources/tcps/\csvcoli_unsmooth}
    }
\end{tabularx}
\caption{Smooth vs. Unsmooth Area Separation. Smooth and unsmooth pixels are filtered using the gradient representations. The images are accompanied by their id in the Painter by Numbers dataset \cite{painter-by-numbers} and their texture complexity. This figure is a validation of our implementation of TCPS as \cite{Peng21TextureRandomization} does not provide sufficient implementation details.}\label{tab:smooth-vs-unsmooth-area}
\end{table*}

\subsection{Photo-metric Distortion}
\begin{figure}
    \subfloat[][]{\centering\includegraphics[width=0.23\linewidth]{Resources/gtr/20078_01}\label{fig:pmd-examples-1}}
    \hfill
    \subfloat[][]{\centering\includegraphics[width=0.23\linewidth]{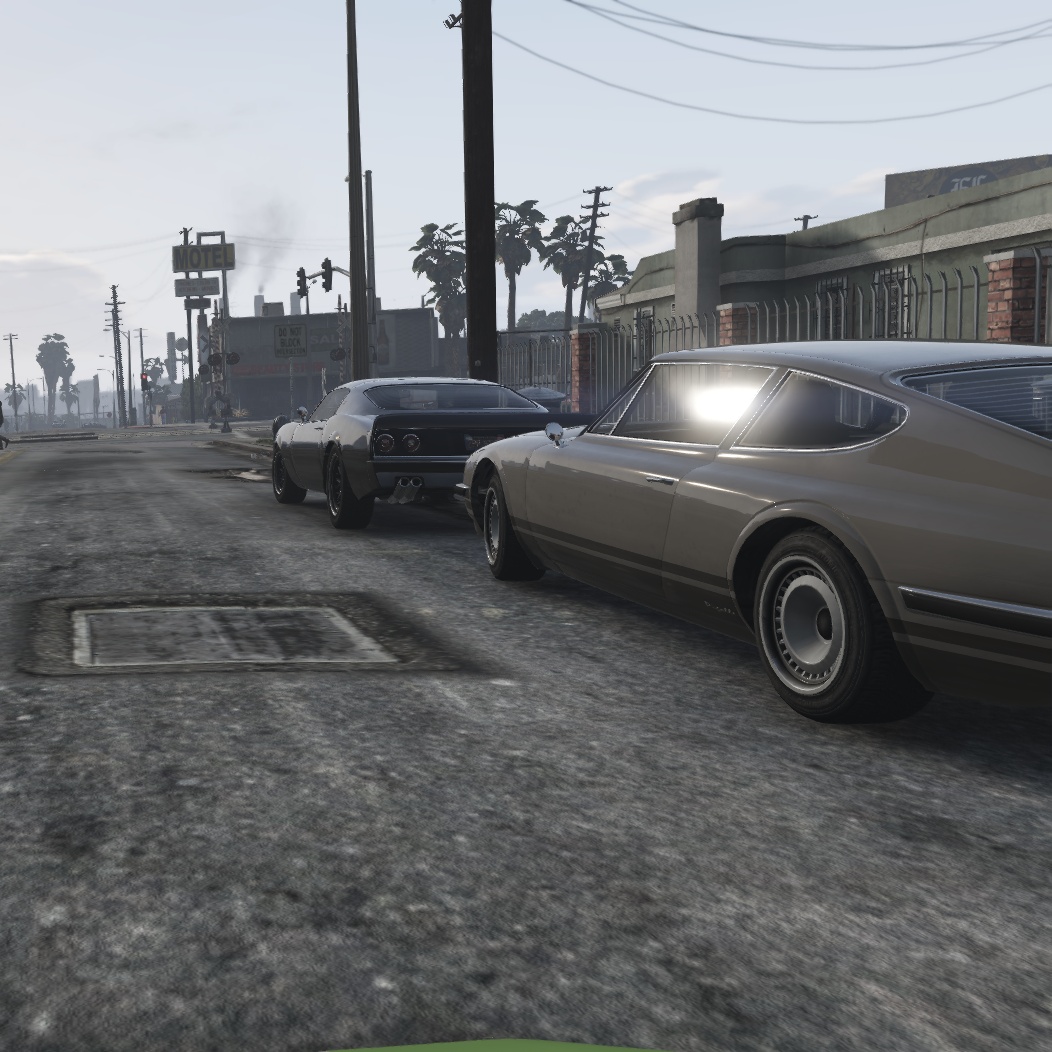}\label{fig:pmd-examples-2}}
    \hfill
    \subfloat[][]{\centering\includegraphics[width=0.23\linewidth]{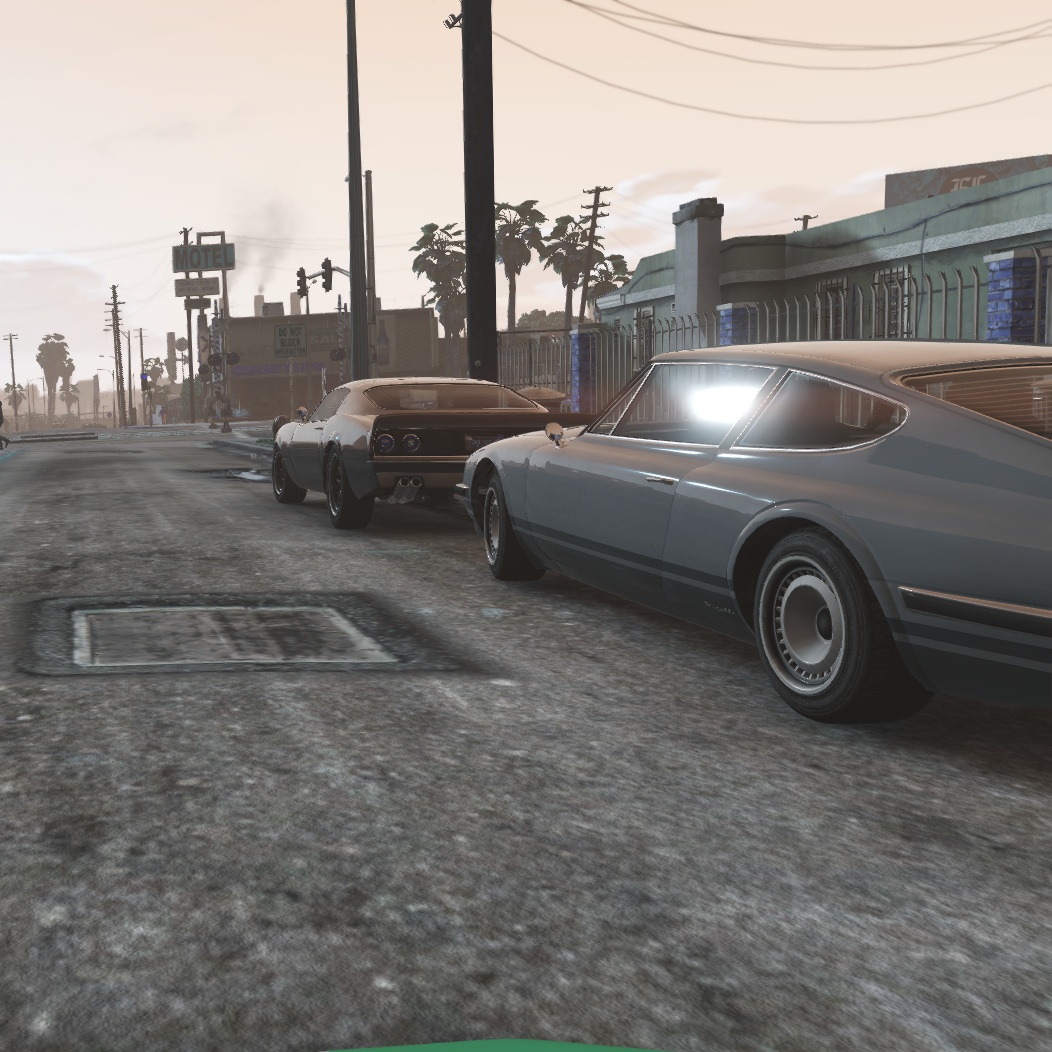}\label{fig:pmd-examples-3}}
    \hfill
    \subfloat[][]{\centering\includegraphics[width=0.23\linewidth]{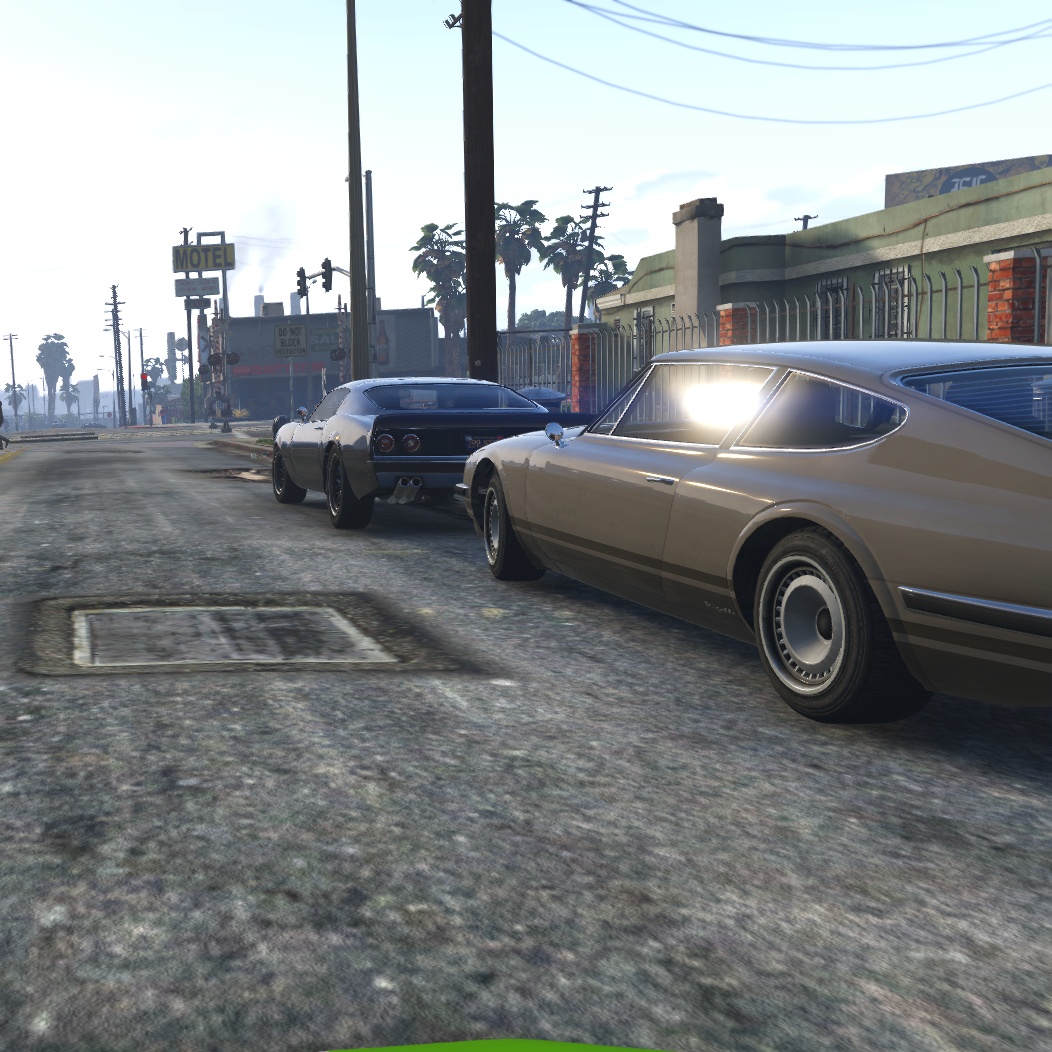}\label{fig:pmd-examples-4}}
    \quad
    \subfloat[][]{\centering\includegraphics[width=0.23\linewidth]{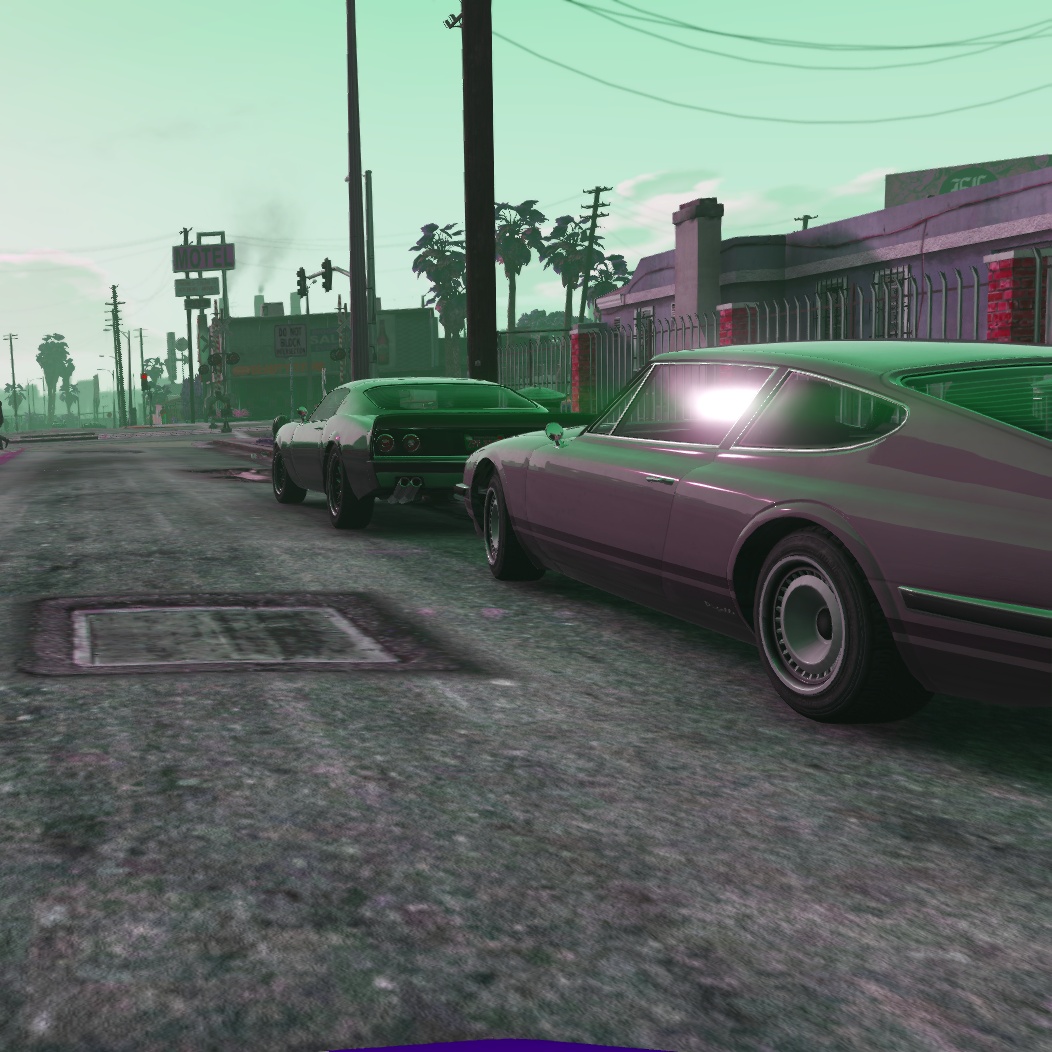}\label{fig:pmd-examples-5}}
    \hfill
    \subfloat[][]{\centering\includegraphics[width=0.23\linewidth]{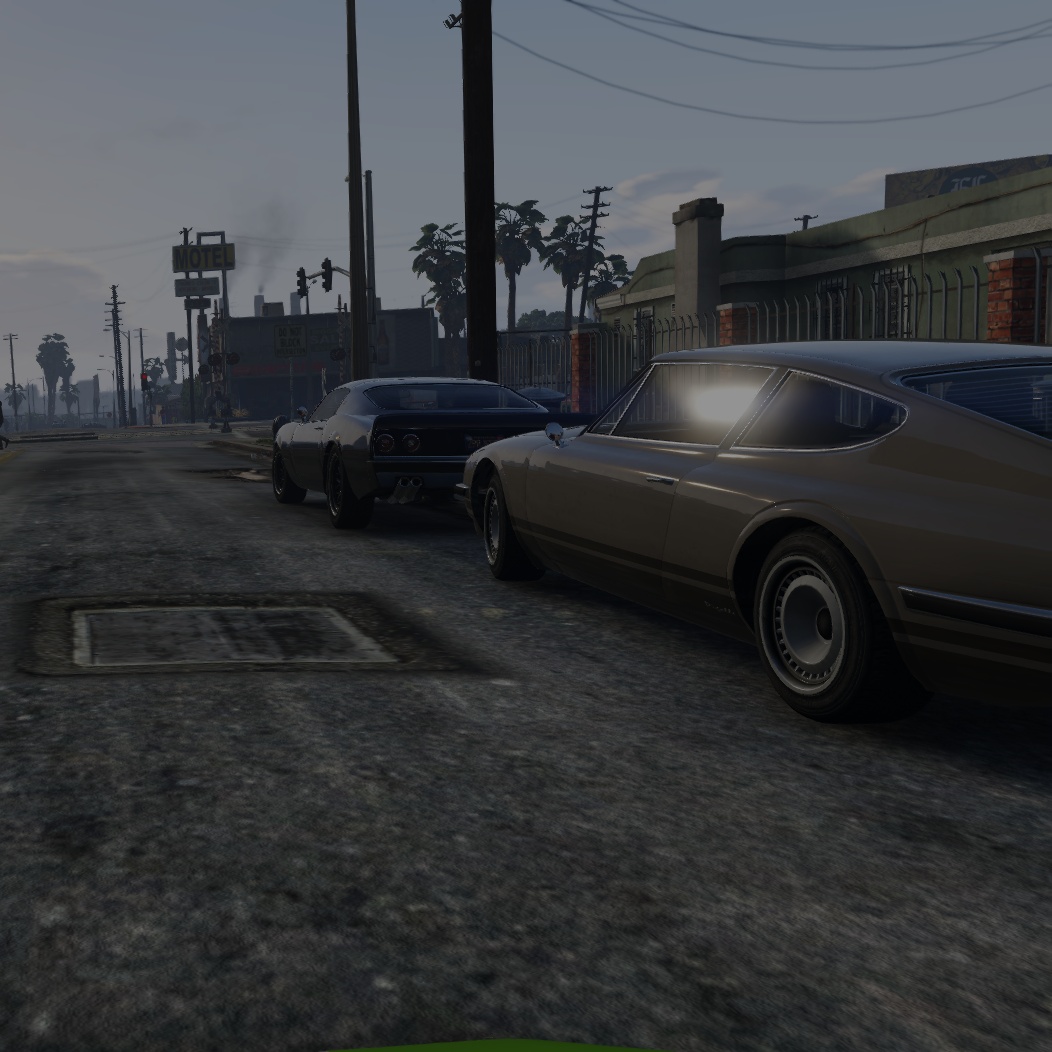}\label{fig:pmd-examples-6}}
    \hfill
    \subfloat[][]{\centering\includegraphics[width=0.23\linewidth]{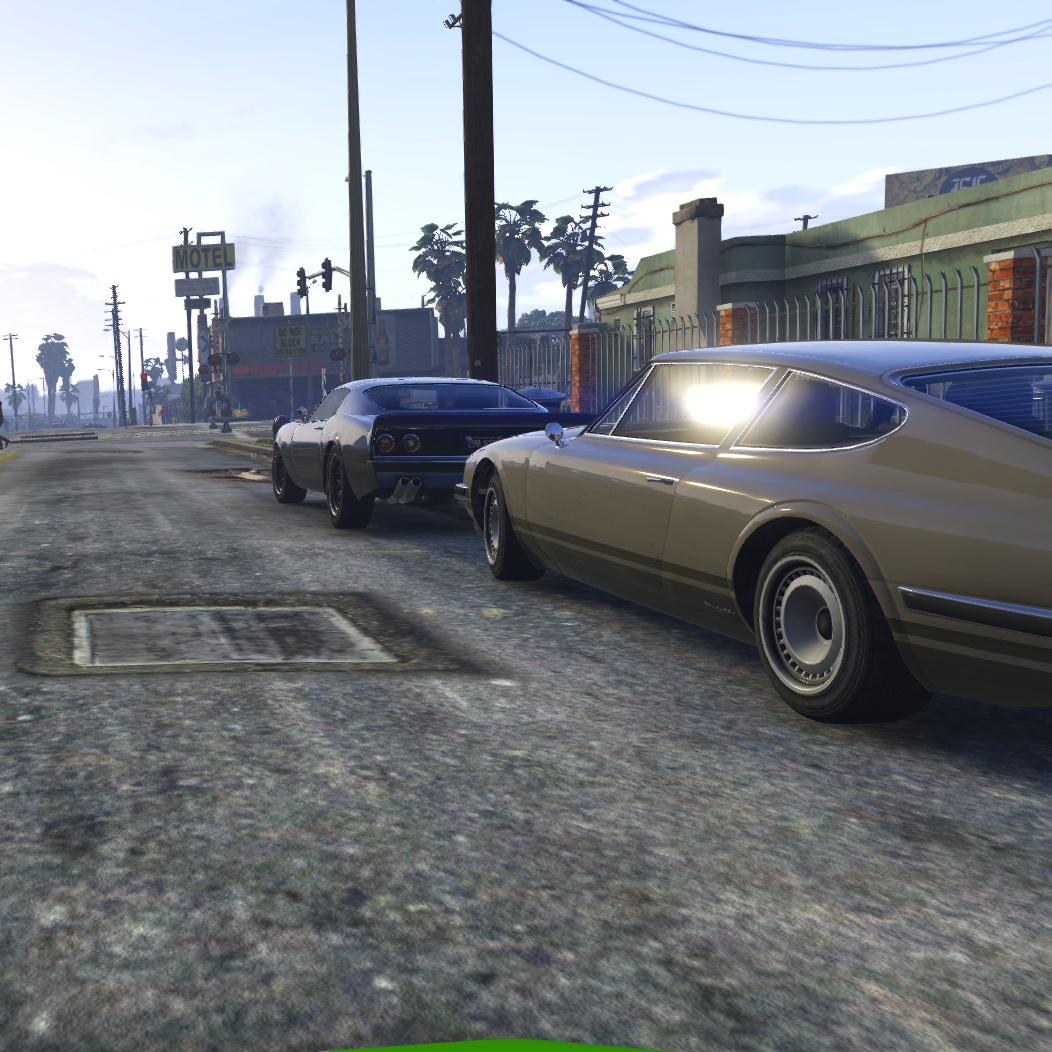}\label{fig:pmd-examples-7}}
    \hfill
    \subfloat[][]{\centering\includegraphics[width=0.23\linewidth]{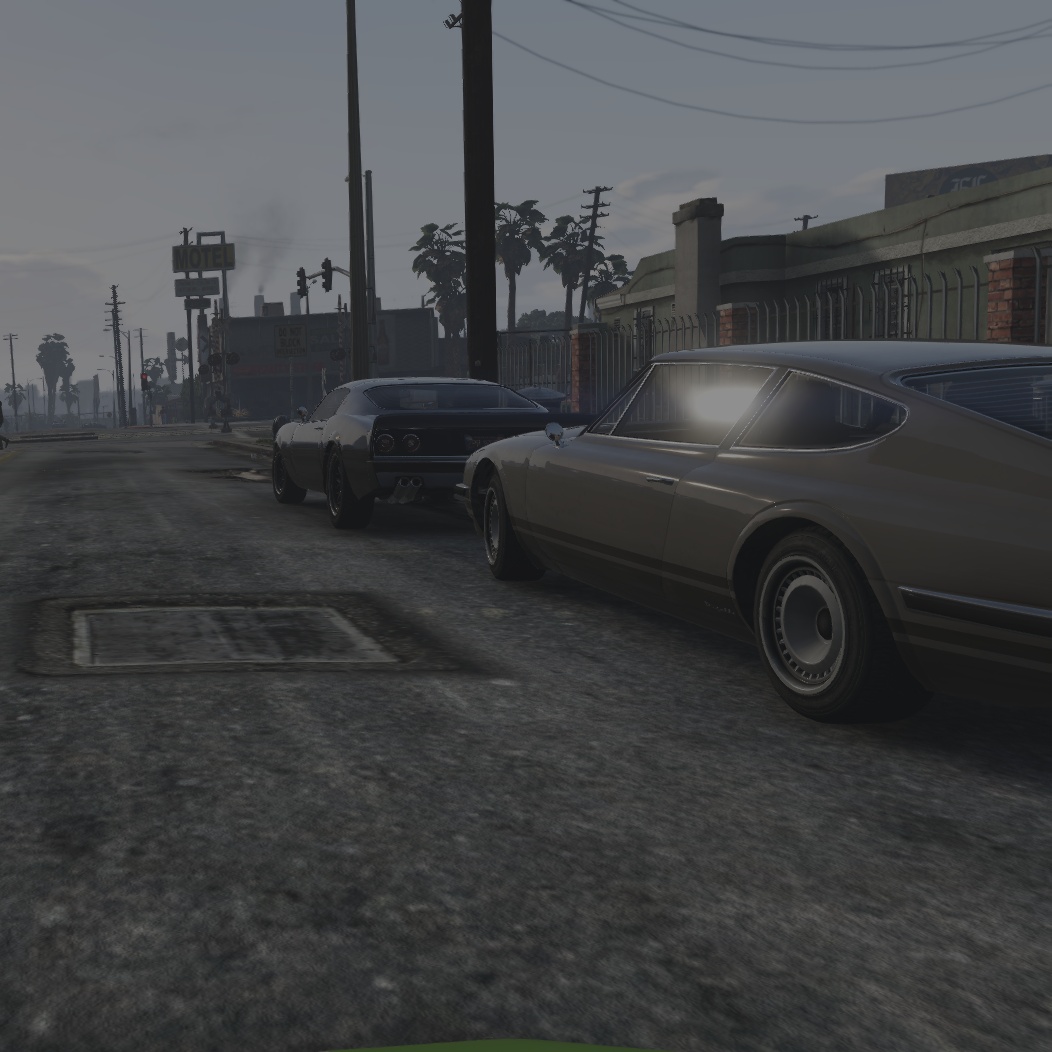}\label{fig:pmd-examples-8}}
    \caption{Examples of Photo-Metric Distortion. (a) shows the source image that is augmented using photo-metric distortion in (b)-(h).}\label{fig:pmd-examples}
\end{figure}

Examples of Photo-metric Distortion are shown in \cref{fig:pmd-examples}

\subsection{Model Architecture}
Even though the DeepLabV2\cite{deeplabv2} architecture is used in related works, there are subtle modifications to it.
For example, it is common to use only the first two out of the four convolutional layers in the segmentation head (cf. \cite{Peng21TextureRandomization,Hoyer22daformer}, which we adopt.
Other works freeze the first batch-normalization layer during training or use the 10-fold learning rate for the segmentation head (cf. \cite{Hoyer22daformer}).
We do not adopt these changes in order to keep the training setup as simple and reproducible as possible, and to avoid introducing confounding factors that could obscure the contribution of the augmentation strategy itself.

\clearpage

\begin{table*}
\begin{center}
\begin{tabularx}{0.85\textwidth}{lcccclcccclc}
    
                    & \multicolumn{4}{c}{R50 Source}     && \multicolumn{4}{c}{R50 StyleMixDG}&\\\cmidrule(lr){2-5}\cmidrule(lr){7-10}
    class           & bdd     & cs    & mv    & mean      && bdd   & cs    & mv    & mean  && +/-\\\midrule
    bicycle         & $01.59$	  & $08.58$ & $19.82$ & $10.00$     && $17.29$ & $20.53$ & $24.82$ & $20.88$ && $+10.88$\\
    building        & $32.88$	  & $46.92$ & $58.33$ & $46.04$     && $65.67$ & $80.18$ & $72.87$ & $72.91$ && $\mathit{+26.86}$\\
    bus             & $04.18$	  & $01.54$ & $07.29$ & $04.34$     && $17.23$ & $27.42$ & $26.92$ & $23.86$ && $+19.52$\\
    car             & $28.08$	  & $\mathit{61.92}$ & $72.13$ & $54.04$     && $73.10$ & $78.36$ & $\mathit{77.23}$ & $76.23$ && $+22.19$\\
    fence           & $12.94$	  & $14.06$ & $21.26$ & $16.09$     && $20.03$ & $16.96$ & $24.80$ & $20.60$ && $+04.51$\\
    m. cycle        & $03.42$	  & $07.58$ & $21.34$ & $10.78$     && $26.33$ & $14.79$ & $28.62$ & $23.25$ && $+12.47$\\
    person          & $25.43$	  & $40.49$ & $49.42$ & $38.45$     && $40.85$ & $59.66$ & $55.88$ & $52.13$ && $+13.68$\\
    pole            & $20.63$	  & $18.17$ & $26.96$ & $21.92$     && $27.87$ & $28.38$ & $31.37$ & $29.21$ && $+07.29$\\
    rider           & $03.35$	  & $07.41$ & $17.71$ & $09.49$     && $17.11$ & $22.37$ & $30.70$ & $23.39$ && $+13.90$\\
    road            & $35.84$	  & $32.46$ & $65.37$ & $44.56$     && $\mathit{78.67}$ & $\mathbf{86.78}$ & $75.21$ & $\mathit{80.22}$ && $\mathbf{+35.66}$\\
    sidewalk        & $10.81$	  & $13.11$ & $26.24$ & $16.72$     && $33.46$ & $42.19$ & $42.65$ & $39.43$ && $+22.71$\\
    sky             & $\mathbf{69.42}$	  & $52.16$ & $\mathbf{76.46}$ & $\mathit{66.01}$     && $\mathbf{80.27}$ & $\mathit{84.26}$ & $\mathbf{81.94}$ & $\mathbf{82.16}$ && $+16.14$\\
    terrain         & $15.76$	  & $09.82$ & $35.33$ & $20.30$     && $28.32$ & $29.97$ & $41.17$ & $33.15$ && $+12.85$\\
    t. light        & $25.74$	  & $23.38$ & $36.55$ & $28.56$     && $31.11$ & $29.51$ & $35.25$ & $31.96$ && $+03.40$\\
    t. sign         & $19.51$	  & $07.25$ & $23.42$ & $16.73$     && $27.29$ & $21.56$ & $35.12$ & $27.99$ && $+11.26$\\
    train           & $00.00$	  & $00.03$ & $10.52$ & $03.52$     && $00.00$ & $02.64$ & $05.68$ & $02.77$ && $-00.74$\\
    truck           & $07.07$	  & $08.35$ & $25.60$ & $13.67$     && $16.34$ & $17.94$ & $26.89$ & $20.39$ && $+06.72$\\
    veget.          & $\mathit{55.39}$	  & $\mathbf{77.87}$ & $\mathit{71.55}$ & $\mathbf{68.27}$     && $68.58$ & $82.85$ & $76.02$ & $75.82$ && $+07.55$\\
    wall            & $02.37$	  & $10.67$ & $06.73$ & $06.59$     && $06.99$ & $24.37$ & $13.49$ & $14.95$ && $+08.36$\\\midrule
    mean            & $19.71$	  & $23.25$ & $35.37$ & $26.11$     && $35.61$ & $40.56$ & $42.45$ & $39.54$ && $+13.43$\\\bottomrule
\end{tabularx}\end{center}
\caption{Detailed Evaluation of StyleMixDG on ResNet50. This table lists the per-class IoU achieved by the ResNet50 backbone using the source-only and StyleMixDG training settings. The best and second-best entries are highlighted in \textbf{bold} and \textit{italics} respectively.}\label{tab:detailed-results-r50}
\end{table*}

\clearpage
\begin{table*}\begin{center}
\begin{tabularx}{0.85\textwidth}{lcccclcccclc}
                    & \multicolumn{4}{c}{R101 Source}     && \multicolumn{4}{c}{R101 StyleMixDG}&\\\cmidrule(lr){2-5}\cmidrule(lr){7-10}
    class           & bdd     & cs    & mv    & mean      && bdd   & cs    & mv    & mean  && +/-\\\midrule
    bicycle         & $09.32$   & $13.69$ & $29.96$ & $17.65$     && $23.11$ & $22.29$ & $28.45$ & $24.62$ && $+06.96$\\
    building        & $44.32$   & $48.54$ & $62.96$ & $51.94$     && $59.70$ & $78.74$ & $61.54$ & $66.66$ && $+14.72$\\
    bus             & $02.89$   & $12.69$ & $08.46$ & $08.01$     && $09.93$ & $25.67$ & $22.96$ & $19.52$ && $+11.51$\\
    car             & $50.18$   & $\mathit{64.83}$ & $73.30$ & $62.77$     && $77.21$ & $83.94$ & $\mathit{82.44}$ & $81.20$ && $+18.43$\\
    fence           & $13.14$   & $13.32$ & $24.99$ & $17.15$     && $19.65$ & $18.99$ & $28.34$ & $22.33$ && $+05.18$\\
    m. cycle        & $14.05$   & $12.86$ & $28.12$ & $18.34$     && $23.24$ & $18.38$ & $35.32$ & $25.65$ && $+07.30$\\
    person          & $32.64$   & $56.98$ & $49.59$ & $46.40$     && $45.30$ & $62.02$ & $59.31$ & $55.54$ && $+09.14$\\
    pole            & $22.44$   & $18.91$ & $30.93$ & $24.09$     && $32.59$ & $34.62$ & $36.61$ & $34.61$ && $+10.51$\\
    rider           & $05.53$   & $13.53$ & $19.88$ & $12.98$     && $12.08$ & $16.26$ & $23.98$ & $17.44$ && $+04.46$\\
    road            & $54.18$   & $44.00$ & $59.99$ & $52.72$     && $\mathbf{82.66}$ & $\mathbf{87.06}$ & $\mathbf{84.00}$ & $\mathbf{84.57}$ && $\mathbf{+31.85}$\\
    sidewalk        & $19.09$   & $23.23$ & $27.88$ & $23.40$     && $35.88$ & $46.80$ & $46.24$ & $42.97$ && $\mathit{+19.57}$\\
    sky             & $\mathbf{71.78}$   & $60.20$ & $\mathbf{79.68}$ & $\mathit{70.55}$     && $\mathit{77.49}$ & $86.58$ & $81.96$ & $\mathit{82.01}$ && $+11.46$\\
    terrain         & $16.98$   & $17.99$ & $33.23$ & $22.73$     && $32.29$ & $33.82$ & $40.04$ & $35.38$ && $+12.65$\\
    t. light        & $22.05$   & $27.53$ & $37.44$ & $29.01$     && $34.65$ & $35.70$ & $39.56$ & $36.64$ && $+07.63$\\
    t. sign         & $22.13$   & $14.26$ & $31.28$ & $22.56$     && $27.94$ & $27.29$ & $41.87$ & $32.37$ && $+09.81$\\
    train           & $00.00$   & $01.39$ & $13.66$ & $05.02$     && $00.00$ & $02.48$ & $10.62$ & $04.37$ && $-00.65$\\
    truck           & $09.02$   & $13.60$ & $30.43$ & $17.68$     && $18.74$ & $23.45$ & $36.47$ & $26.22$ && $+08.54$\\
    veget.          & $\mathit{60.56}$   & $\mathbf{80.00}$ & $\mathit{73.44}$ & $\mathbf{71.33}$     && $71.83$ & $\mathit{85.05}$ & $76.49$ & $77.79$ && $+06.46$\\
    wall            & $01.40$   & $07.72$ & $05.32$ & $04.81$     && $06.15$ & $22.90$ & $21.60$ & $16.88$ && $+12.07$\\\midrule
    mean            & $24.83$   & $28.70$ & $37.92$ & $30.48$     && $36.34$ & $42.74$ & $45.15$ & $41.41$ && $+10.93$\\\bottomrule
\end{tabularx}\end{center}
\caption{Detailed Evaluation of StyleMixDG on ResNet101. This table lists the per-class IoU achieved by the ResNet101 backbone using the source-only and StyleMixDG training settings. The best and second-best entries are highlighted in \textbf{bold} and \textit{italics} respectively.}\label{tab:detailed-results-r101}
\end{table*}

\end{document}